\documentclass[10pt,journal,compsoc]{IEEEtran}

\ifCLASSOPTIONcompsoc
\usepackage[nocompress]{cite}
\else
\usepackage{cite}
\fi
\usepackage{epsfig}
\usepackage{array}
\usepackage{xcolor}
\usepackage{url}
\usepackage{bm}
\usepackage{tabularx}
\usepackage{color}
\usepackage{balance}
\usepackage{mathtools}
\usepackage{pifont}
\usepackage{booktabs}       
\usepackage{tikz}
\usepackage{epstopdf}
\usepackage{lineno}

\usepackage{epsfig}
\usepackage{amsmath}
\usepackage{amssymb}
\usepackage{helvet}
\usepackage{courier}
\usepackage{multirow}
\usepackage{amsopn}
\usepackage{graphicx}
\usepackage{array}
\usepackage{xcolor}
\usepackage{tabularx}
\usepackage[ruled,vlined]{algorithm2e}
\usepackage[figuresright]{rotating}
\usepackage{indentfirst}

\usepackage{times}  
\usepackage{helvet}  
\usepackage{courier}  
\usepackage{url}  
\usepackage{amsthm}
\usepackage{setspace}
\usepackage{bm}
\usepackage[figuresright]{rotating}
\usepackage{arydshln}
\usepackage{float}
\usepackage{csquotes}
\usepackage{makecell}
\usepackage{microtype}      
\usepackage{algpseudocode}  
\usepackage{algorithmicx}   
\usepackage{threeparttable}

\makeatletter
\newcommand*\bigcdot{\mathpalette\bigcdot@{.5}}
\newcommand*\bigcdot@[2]{\mathbin{\vcenter{\hbox{\scalebox{#2}{$\m@th#1\bullet$}}}}}
\makeatother

\newcommand{\R}{\mathbb{R}}
\newcommand{\E}{\mathbb{E}}

\newcommand{\cmark}{\ding{51}}%
\newcommand{\xmark}{\ding{55}}%

\newcommand{\ra}[1]{\renewcommand{\arraystretch}{#1}}
\newcommand{\co}[1]{\setlength\tabcolsep{#1}}
\newcommand{\bftab}{\fontseries{b}\selectfont}

\newcommand{\smallcite}[1]{}

\newcommand{\seen}{\textit{seen}}

\newcommand{\x}{\bm{x}}
\newcommand{\y}{\bm{y}}
\newcommand{\z}{\bm{z}}

\newtheorem{lemma}{Lemma}[section]

\makeatletter
\newcommand{\thickhline}{%
	\noalign {\ifnum 0=`}\fi \hrule height 1pt
	\futurelet \reserved@a \@xhline
}
\makeatother

\begin{document}
	
	\title{Decoding Visual Neural Representations by Multimodal Learning of Brain-Visual-Linguistic Features}

	\author{Changde~Du,
		Kaicheng~Fu,
		Jinpeng~Li,
		and~Huiguang~He,~\IEEEmembership{Senior Member,~IEEE}
		
		\IEEEcompsocitemizethanks{
			\IEEEcompsocthanksitem C. Du, K. Fu and H. He are with the Research Center for Brain-Inspired Intelligence, State Key Laboratory of Multimodal Artificial Intelligence Systems, Institute of Automation, Chinese Academy of Sciences, Beijing 100190, China. K. Fu and H. He are also with the School of Artificial Intelligence, University of Chinese Academy of Sciences (UCAS), Beijing 100049, China (e-mail: changde.du@ia.ac.cn, fukaicheng2019@ia.ac.cn, huiguang.he@ia.ac.cn).
			\IEEEcompsocthanksitem J. Li is with the Ningbo HwaMei Hospital, UCAS, Zhejiang 315010, China (e-mail: lijinpeng@ucas.ac.cn).
		}
		\thanks{This work was supported in part by the National Key R\&D Program of China 2022ZD0116500; in part by the National Natural Science Foundation of China under Grant 62206284, Grant 62020106015 and Grant 61976209; in part by the
			Strategic Priority Research Program of Chinese Academy of Sciences under Grant XDB32040000; and in part by the CAAI-Huawei MindSpore Open Fund. (Corresponding author: Huiguang He)}
	}


	\IEEEcompsoctitleabstractindextext{%
		\begin{abstract}  Decoding human visual neural representations is a challenging task with great scientific significance in revealing vision-processing mechanisms and developing brain-like intelligent machines. Most existing methods are difficult to generalize to novel categories that have no corresponding neural data for training. The two main reasons are 1) the under-exploitation of the multimodal semantic knowledge underlying the neural data and 2) the small number of paired (\emph{stimuli-responses}) training data. To overcome these limitations, this paper presents a generic neural decoding method called BraVL that uses multimodal learning of brain-visual-linguistic features. We focus on modeling the relationships between brain, visual and linguistic features via multimodal deep generative models. Specifically, we leverage the mixture-of-product-of-experts formulation to infer a latent code that enables a coherent joint generation of all three modalities. To learn a more consistent joint representation and improve the data efficiency in the case of limited brain activity data, we exploit both intra- and inter-modality mutual information maximization regularization terms.  In particular, our BraVL model can be trained under various semi-supervised scenarios to incorporate the visual and textual features obtained from the extra categories.  Finally, we construct three trimodal matching datasets, and the extensive experiments lead to some interesting conclusions and cognitive insights: 1) decoding novel visual categories from human brain activity is practically possible with good accuracy; 2) decoding models using the combination of visual and linguistic features perform much better than those using either of them alone; 3) visual perception may be accompanied by linguistic influences to represent the semantics of visual stimuli. Code and data: {\color{red} https://github.com/ChangdeDu/BraVL}.				
		\end{abstract}
		\begin{IEEEkeywords}
			Generic neural decoding, brain-visual-linguistic embedding, multimodal Learning, mutual information maximization
	\end{IEEEkeywords}}

	\maketitle

	\IEEEdisplaynontitleabstractindextext
	\IEEEpeerreviewmaketitle

	\IEEEraisesectionheading{\section{Introduction}\label{sec:introduction}}
	\IEEEPARstart{H}uman visual capabilities are superior to current artificial systems. Many cognitive neuroscientists and artificial intelligence researchers have been committed to reverse-engineering the human mind, to decipher and simulate the mechanism of the brain and to promote the development of brain-inspired computational models \cite{kamitani2005decoding,du2018reconstructing,palazzo2020decoding}. Although there is an increasing interest in the visual neural representation decoding task, inferring visual category from human brain activity for novel classes remains the boundary to explore. Zero-Shot Neural Decoding (ZSND) based on functional Magnetic Resonance Imaging (fMRI) or electroencephalography (EEG) data aims to tackle this problem \cite{mitchell2008predicting,NIPS2009_1543843a,horikawa2017generic}. In the ZSND, we have access to a set of brain activity of seen classes, and the objective is to leverage the visual \cite{kay2008identifying,horikawa2017generic} or linguistic \cite{mitchell2008predicting,NIPS2009_1543843a} semantic knowledge to learn a generic neural decoder that enables generalization to novel classes at test time. These studies not only help to reveal the cognitive mechanism of the human brain, but also provide a technical basis for the development of Brain-Computer Interfaces (BCIs).
	
	\begin{figure}[t!]
		\centering 
		\includegraphics[scale=0.6]{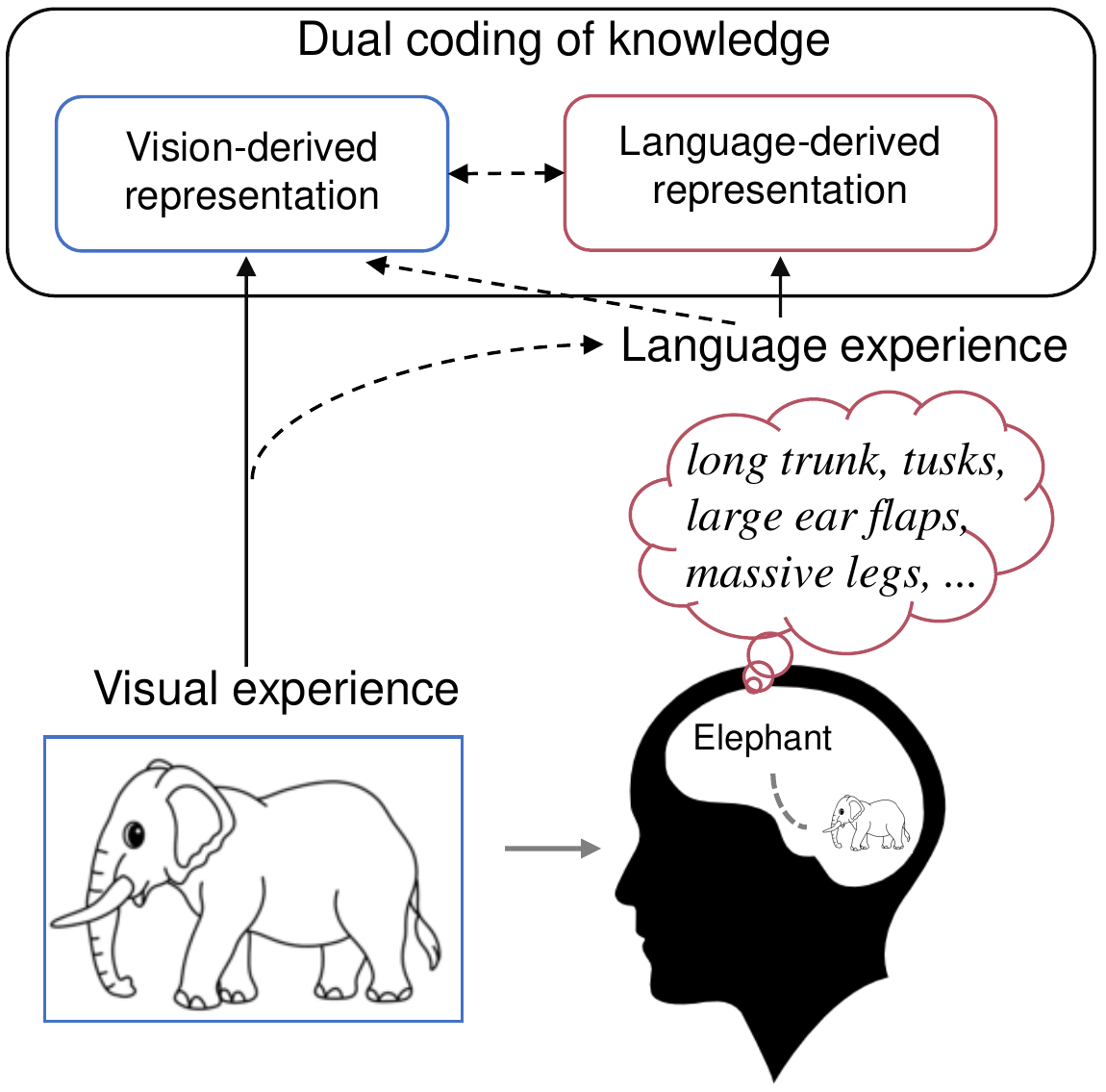}
		\vspace{-6pt}
		\caption{
			Dual coding of knowledge in the human brain. When we see a picture of elephant, we will spontaneously retrieve the knowledge of elephant in our mind. Then, the concept of elephant is encoded in the brain both visually and linguistically, where language, as a valid prior experience, contributes to shaping vision-derived representations. 
		}
		\vspace{-16pt}
		\label{fig:dualcoding}
	\end{figure}

	Existing visual neural representation decoding methods mostly resort to visual semantic knowledge, such as features extracted from viewed images on the basis of Gabor wavelet filters \cite{kay2008identifying} or a Convolutional Neural Network (CNN) \cite{horikawa2017generic,o2019zero,du2020structured}. 
	However, the human ability to detect, discriminate, and recognize perceptual visual stimuli is influenced by both visual features and people's prior experiences \cite{LUPYAN2020930}.
	For example, when we see a familiar object, we spontaneously retrieve the knowledge of that object and the entity relationships that object forms in our mind. As shown in Fig. \ref{fig:dualcoding}, cognitive neuroscience research on dual-coding theory  \cite{paivio2013imagery,bi2021dual} also considers concrete concepts to be encoded in the brain both visually and linguistically, where language, as a valid prior experience, contributes to shaping vision-derived representations. Moreover, in brain-inspired computational modeling, large-scale multimodal pretrained models \cite{radford2021learning,fei2022towards} formed by combining image and text representations provide a better proxy for human-like intelligence. Therefore, we argue that the recorded brain activity should be decoded using a combination of not only the visual semantic features that were in fact presented as clues, but also a far richer set of linguistic semantic features typically related to the target object.

\begin{figure*}[t!]
		\centering 
		\includegraphics[scale=0.563]{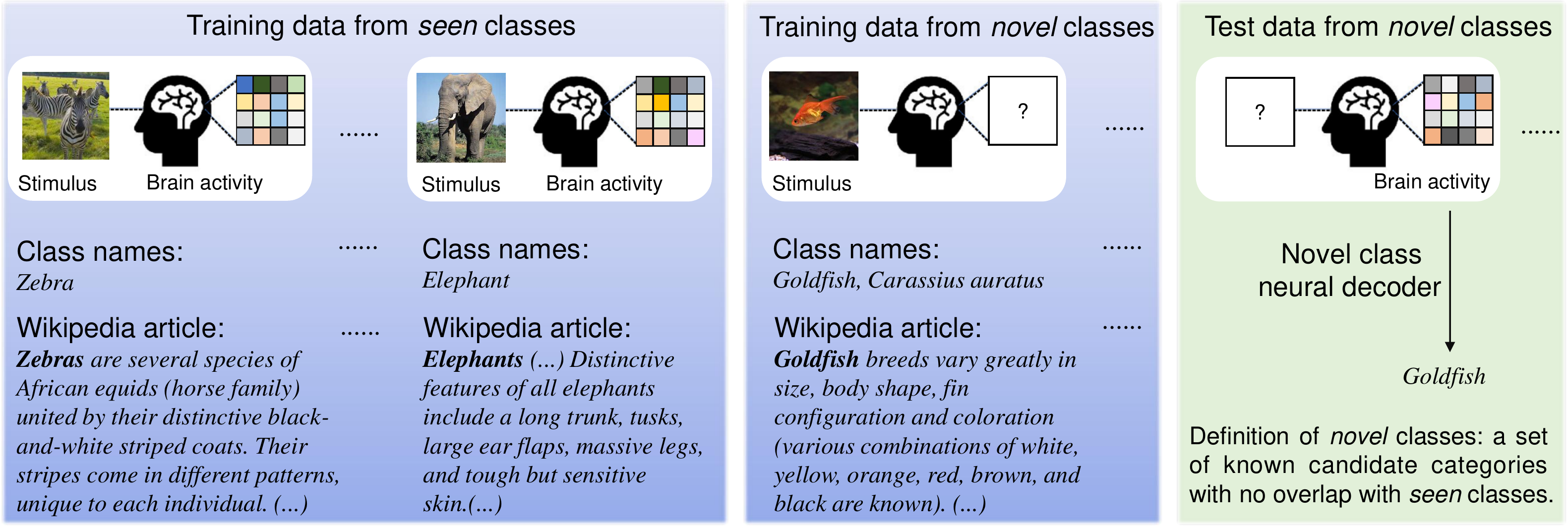}
		\vspace{-7pt}
		\caption{Image stimuli, evoked brain activity and their corresponding textual data.  We can only collect brain activity for a few categories, but we can easily collect images and/or text data for almost all categories. Therefore, for seen classes, we assume that the brain activity, visual images and corresponding textual descriptions are available for training, whereas for novel classes, only visual images and textual descriptions are available for training. The test data are brain activity from the novel classes.
		}
		\vspace{-8pt}
		\label{fig:data}
	\end{figure*}

	Although several studies have addressed the idea of decoding naturalistic visual experiences from brain activity using purely linguistic features \cite{nishida2018decoding,vodrahalli2018mapping}, they merely use standard word vectors of class names that are automatically extracted from large corpora such as Common Crawl. Actually, the word vectors of class names are barely aligned with visual information \cite{bujwid-sullivan-2021-large}. As a result, the neural decoding accuracy is still far from the practical criterion. Is it possible to build a language representation that is more consistent with visual cognition, with richer visual semantics?  Previous studies using Wikipedia text descriptions to represent image classes have shown some positive signs \cite{bujwid-sullivan-2021-large,kil2021revisiting}.  For example, as shown in Fig. \ref{fig:data}, the page ``Elephants" contains phrases ``long trunk, tusks, large ear flaps, massive legs" and ``tough but sensitive skin" that exactly match the visual attributes. Intuitively, Wikipedia articles capture richer visual semantic information than class names. Here, we argue that using natural languages such as Wikipedia articles as class descriptions will yield better neural decoding performance than using class names.

	Motivated by the aforementioned discussions, we proposed a biologically plausible neural decoding method, called BraVL, to infer novel image categories from human brain activity by the joint learning of brain-visual-linguistic features.  Our model focuses on modeling the relationships between brain activity and multimodal semantic knowledge, i.e., visual semantic knowledge extracted from images and textual semantic knowledge obtained from rich Wikipedia descriptions of classes. Specifically, we developed a multimodal auto-encoding variational Bayesian learning framework, in which we used the mixture-of-product-of-experts formulation \cite{sutter2021generalized} to infer a latent code that enables coherent joint generation of all three modalities. To learn a more consistent joint representation and improve the data efficiency in the case of limited brain activity data, we further introduced both the intra- and inter-modality Mutual Information (MI) regularization terms. In particular, our BraVL model can be trained under various semi-supervised learning scenarios to incorporate the extra visual and textual features obtained from the large-scale image categories in addition to the image categories of training data. Furthermore, we collected the corresponding textual descriptions for two popular Image-fMRI datasets \cite{horikawa2017generic,shen2019deep} and one Image-EEG dataset \cite{gifford2022large}, hence forming three new trimodal matching (brain-visual-linguistic) datasets. The experimental results give us three significant observations. First, models using the combination of visual and textual features perform much better than those using either of them alone. Second, using natural languages as class descriptions yields higher neural decoding performance than using class names.  Third,  either unimodal or bimodal extra data can remarkably improve decoding accuracy.
	
	\textbf{Contributions.}\ \	
	In summary, our main contributions are listed as follows: 1) We combine visual and linguistic knowledge for neural decoding of visual categories from human brain activity for the first time. 2) We develop a new multimodal learning model with specially designed intra- and inter-modality MI regularizers to achieve more consistent brain-visual-linguistic joint representations and improved data efficiency. 3) We contribute three trimodal matching datasets, containing high-quality brain activity, visual features and textual features. Our code and datasets have been released\footnote{https://github.com/ChangdeDu/BraVL} to facilitate further research. 4) Our experimental results show several interesting conclusions and cognitive insights about the human visual system. 
	
	\section{Related work}
	\textbf{Neural decoding of visual categories.}
	Estimating the semantic categories of viewed images from evoked brain activity has long been a sought objective.  Previous works mostly relied on a classification-based approach, where a classifier is trained to build the relationship between brain activity and the predefined labels using fMRI \cite{haxby2001distributed,kamitani2005decoding,van2010efficient,damarla2013decoding}  or EEG \cite{ahmed2021object,palazzo2020decoding,spampinato2017deep,du2021multimodal} data.  However, this kind of method is restricted to the decoding of a specified set of categories. 
	To allow novel category decoding, several identification-based methods \cite{kay2008identifying,horikawa2017generic,akamatsu2020brain} were proposed by characterizing the relationship between brain activity and
	visual semantic knowledge, such as image features extracted from  Gabor wavelet filters \cite{kay2008identifying} or a CNN
	\cite{horikawa2017generic,akamatsu2020brain}. Although these methods allow the identification of a large set of possible image categories, the decoding accuracy significantly depends on the large number of paired \emph{stimuli-responses} data, which is difficult to collect. Therefore, accurately decoding novel image categories remains a challenge. 
	Neurolinguistic studies have shown that distributed word representations are also correlated with evoked brain activity \cite{huth2016natural,pereira2018toward,nishida2018decoding,kivisaari2019reconstructing}. Encouraged by these findings, we associate brain activity with multimodal semantic knowledge, i.e., not only visual features but also textual features. 
	In particular, rather than learning a mapping directly to multimodal semantic knowledge, we focus on creating a latent space that could describe any valid categories, and then learn a mapping between brain activity and this latent space.
	
	\textbf{Zero-shot learning (ZSL).}
	ZSL is a classification problem where the label space is divided into two distinct sets: seen and novel classes \cite{revise,good_bad,latem}. To alleviate the problem of \seen-\emph{novel}~domain shift, training samples typically consist of semantic knowledge such as attributes~\cite{revise,cada} or word embeddings~\cite{devise} that bridge the semantic gap between seen and novel classes. Semantic knowledge of these types reflects human heuristics, and can therefore be  extended and transferred from \seen~classes to \emph{novel} ones, specifying the semantic space in ZSL. 
	ZSL methods can be roughly divided into three categories, depending on the method used to inject semantic knowledge: 1) learning \emph{instance$\rightarrow$semantic} projections~\cite{devise,eszsl}, 2) learning \emph{semantic$\rightarrow$instance} projections~\cite{verma2018generalized,zhang2019co}, and 3) learning the projections of instance and semantic spaces to a shared latent space \cite{zhang2015zero,cada}. Our approach falls into the third category. Recently, ZSL researchers have achieved success through the use of deep generative models \cite{cada,f-vaegan-d2}, which are used for synthesizing data features as a data augmentation mechanism in ZSL. In our work, we use the ZSL paradigm to solve the novel class neural decoding task. Although visual and linguistic semantic knowledge are observable for the novel class, there was no brain activity data. Therefore, the novel class neural decoding can be regarded as a zero-shot classification problem.
	
	\textbf{Multimodal learning.}
	Multimodal learning is inspired by cognitive science research, suggesting that human semantic knowledge relies on perceptual and sensori-motor experience. 
	Multimodal semantic models using both linguistic representations and visual perceptual information have been proven successful in a range
	of Natural Language Processing (NLP) tasks, such as learning word embeddings \cite{hill2014learning,radford2021learning}.
	Several studies have addressed the idea of decoding linguistic nouns from brain activity using both linguistic and visual perceptual information \cite{anderson2017visually,bulat2017speaking,davis2019deconstructing}. Anderson et al. applied linguistic and visually-grounded computational models to decode the neural representations of a set of concrete and abstract nouns \cite{anderson2017visually}. Davis et al. constructed multimodal models combining linguistic and three kinds of visual features, and evaluated the models on the task of decoding brain activity associated with the meanings of nouns \cite{davis2019deconstructing}.
	In contrast to the above studies that have leveraged visual features to boost the neural decoding of linguistic nouns, we introduce textual features to enhance the neural decoding of visual categories.

	\textbf{Mutual information maximization.}
	For two random variables $X$ and $Y$ whose joint probability distribution is $p(x, y)$, the mutual information (MI) between them is defined as $I(X; Y)=\E_{p(x,y)}\left[\log\frac{p(x,y)}{p(x)p(y)}\right]$.
	Furthermore, as a Shannon entropy-based quantity, MI can also be written as $I(X; Y) = H(X) - H(X|Y)$, where $H(X)$ is the Shannon entropy and $H(X|Y)$ is the conditional entropy. As a pioneer, \cite{alemi2016deep}~first incorporated MI-related optimization into deep learning. Since then, many works have demonstrated the benefit of the MI-maximization in deep representation learning~\cite{bachman2019learning,he2020momentum, amjad2019learning}. Since directly optimizing MI in high-dimensional spaces is nearly impossible, many approximation methods with variational bounds have been proposed~\cite{belghazi2018mutual, cheng2020club, poole2019variational}. In our work, we apply MI-maximization at both the intra- and inter-modality levels in multimodal representation learning. We prove that inter-modality MI-maximization is equivalent to multimodal contrast learning.

	\section{Multimodal learning of brain-visual-linguistic features}
	
	\subsection{Problem definition}
	In real world applications, we can only collect brain activity for a few visual categories, but we can easily collect images and/or text data for almost all categories. If we can make full use of the plentiful image and text data without corresponding brain activity, we have opportunities to improve the generalization performance of neural decoding models. 
	Therefore, as shown in Fig. \ref{fig:data}, we assume that brain activity, visual images (from ImageNet) and class-specific textual descriptions (from Wikipedia) are provided for the \emph{seen} classes, but only visual and textual information are provided for \emph{novel/unseen} classes . Our goal is to learn a classifier (i.e., neural decoder) that can classify \emph{novel} class brain activity at test time. Note that the \emph{novel} classes are a set of known candidate categories with no overlap with the seen classes (rather than the infinite  arbitrary categories).
	
	Let $\mathcal{D}^{seen}=\{(\x_b, \x_v, \x_t, \y)|\x_b \in X_b^{s}, \x_v \in X_v^{s}, \x_t \in X_t^{s}, \y \in Y^{s} \}$  be the set of \emph{seen} class data, where $X_b^{s}$ corresponds to the set of brain activity (fMRI) features, $X_v^{s}$ denotes the visual features, $X_t^{s}$ denotes the textual features and $Y^{s}$ denotes the set of \emph{seen} class labels. Similarly, the data for \emph{novel/unseen} classes are defined as $\mathcal{D}^{novel}=\{(\x_v^n, \x_t^n, \y^n)|\x_v^n \in X_v^{n}, \x_t^n \in X_t^{n}, \y^n \in Y^{n}\}$, where $X_v^{n}$, $X_t^{n}$ and $Y^{n}$ denote the visual features, textual features and class labels of the \emph{novel} classes, respectively. The \emph{seen} class labels  $Y^{s}$ and  the \emph{novel} class labels $Y^{n}$ are disjointed in categories, i.e., $Y^{s} \cap Y^{n} = \emptyset$. Note that the \emph{novel} class brain activity data $X_b^{n}$ is unavailable during model training, and it will only be used at test time. 
	
	Let $b, v$ and $t$ represent the subscripts of the brain, visual, and textual modality, respectively. For any given modality subscript $m$ ($m\in \{b, v, t\}$), the unimodal feature matrix $X_m\in \R^{N_m\times d_m}$, where $X_m = X_m^{s} \bigcup X_m^{n}$, $N_m = N_m^s + N_m^n$ is the sample size and $d_m$ is the feature dimension of modality $m$.  
	
	\vspace{-1pt}
	\subsection{Brain, image and text  preprocessing} \label{Modality_Encoding}
\begin{figure}[htbp!]
		\centering 
		\includegraphics[scale=0.90]{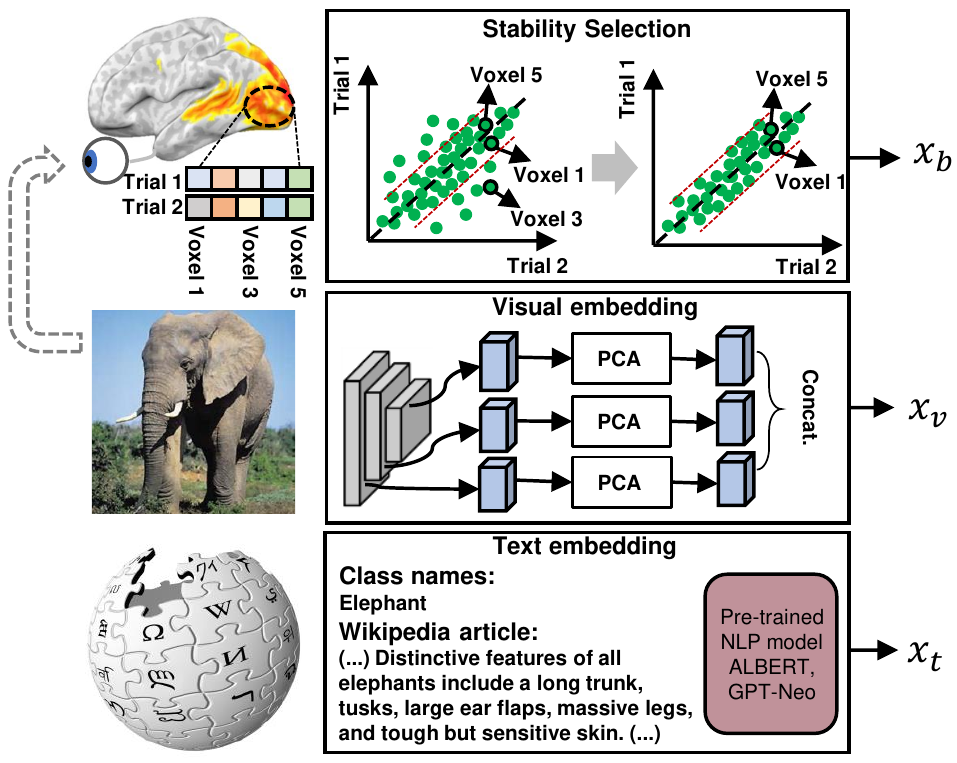}
		\vspace{-7pt}
		\caption{
			Data preprocessing. We preprocess the raw inputs into feature representations with modality-specific feature extractors. 
		}
		\vspace{-7pt}
		\label{fig:preprocess}
	\end{figure}

	As shown in Fig. \ref{fig:preprocess}, we first preprocess the raw inputs into feature representations with modality-specific feature extractors.

	\textbf{Stability selection of brain voxels.} 
	Brain activity differs from trial to trial, even for an identical visual stimulus. To improve the stability of neural decoding, we used stability selection for fMRI data, in which the voxels showing the highest consistency in activation patterns across distinct trials for an identical visual stimulus were selected for the analysis, following \cite{kivisaari2019reconstructing}. 
	This stability is quantified for each voxel as the mean Pearson correlation coefficient across all pairwise combinations of the trials. 
	In particular, the stable voxel was selected separately on each brain region to avoid the selected voxels concentrated in the local brain region to ensure that a portion of high-quality voxels were retained in each brain region. This operation can effectively reduce the dimension of fMRI data and suppress the interference caused by noisy voxels without seriously affecting the discriminative ability of brain features.
	For each selected brain voxel, its response vector to the visual stimuli belonging to the \emph{seen} classes is normalized (across stimuli, zero-mean and unit-variance). Note that we used only the training fMRI data belonging to the \emph{seen} classes to calculate the normalization parameters (i.e., the mean and variance of each voxel) of each selected voxel, and the calculated mean and variance were used to normalize both the training and testing fMRI data separately. After stability selection and normalization, we perform Principal Component Analysis (PCA) on the training fMRI data belonging to the \emph{seen} classes for dimensionality reduction. The brain feature dimensions after keeping 99\% of the variance using PCA are shown in Section 4.1. Note that the test samples are not included in the PCA fitting, and we use only the training samples to estimate the PCA mapping weights. After PCA fitting, the estimated mapping weights are directly applied to the test samples to obtain the dimension-reduced test samples.
	
	\textbf{Feature extraction of visual images.}
	We use a powerful VGG-style ConvNet, referred to as RepVGG \cite{ding2021repvgg}, to extract hierarchical visual features from the images.  Specifically, we use the Timm library\footnote{https://github.com/rwightman/pytorch-image-models} to extract the intermediate feature maps with different strides in the RepVGG-b3g4 model, which had been pretrained to achieve 80.21\% top-1 accuracy on ImageNet \cite{ding2021repvgg}.
	Similar to the brain feature processing pipeline, the extracted visual features of \emph{seen} classes are flattened and normalized first, and then dimensionality reduction is performed using PCA to keep 99\% of the variance. 
	
	\textbf{Embedding of textual descriptions.}
	In early studies of language processing and understanding, generating vectors to represent sentences is typically done by averaging vectors for the content words \cite{mitchell2010composition}. This method of obtaining sentence vector by average pooling of word vectors has been successfully applied in many linguistic neural encoding and decoding studies \cite{schrimpf2021the, pereira2018toward}, and has achieved impressive decoding results. With the development of NLP method, researchers started to input individual sentence into Transformer-based models~\cite{vaswani2017attention}, such as BERT~\cite{DevlinCLT19}, and to derive fixed-size sentence embeddings, which have been found to be very effective for neural encoding \cite{schrimpf2021the}. To obtain sentence embedding from BERT-like NLP models, the most commonly used approach is to average the output layer (known as token embeddings) or by using the output of the first token (the [CLS] token). As shown in a previous linguistic neural decoding study \cite{gauthier2019linking}, these two common practices yield similar qualitative results.
	Here, we use ALBERT~\cite{lan2019albert} and GPT-Neo \cite{gpt-neo} as text encoders, and we use the mean of token embeddings as the sentence embedding.\footnote{https://github.com/huggingface/transformers} 
	
	Due to the constraint on the input sequence length for ALBERT and GPT-Neo, we cannot directly input the entire Wikipedia article into the model. To encode articles that can be longer than the maximal length, we alternatively split the article text into partially overlapping sequences of 256 tokens with an overlap of 50 tokens. Concatenating multiple sentence embeddings will lead to an undesirable `curse of dimensionality' issue. Therefore, we use the average-pooled representation of multiple sequences to encode the entire article. This average-pooling strategy has also been successfully used in a recent linguistic neural encoding study \cite{oota-etal-2022-neural}. Similarly, if a class has multiple corresponding articles in Wikipedia, we average the representations obtained from each of them. See Appendix for the degree of heterogeneity of text features under average-pooling.

	\begin{figure*}[t!]
		\centering 
		\includegraphics[scale=0.745]{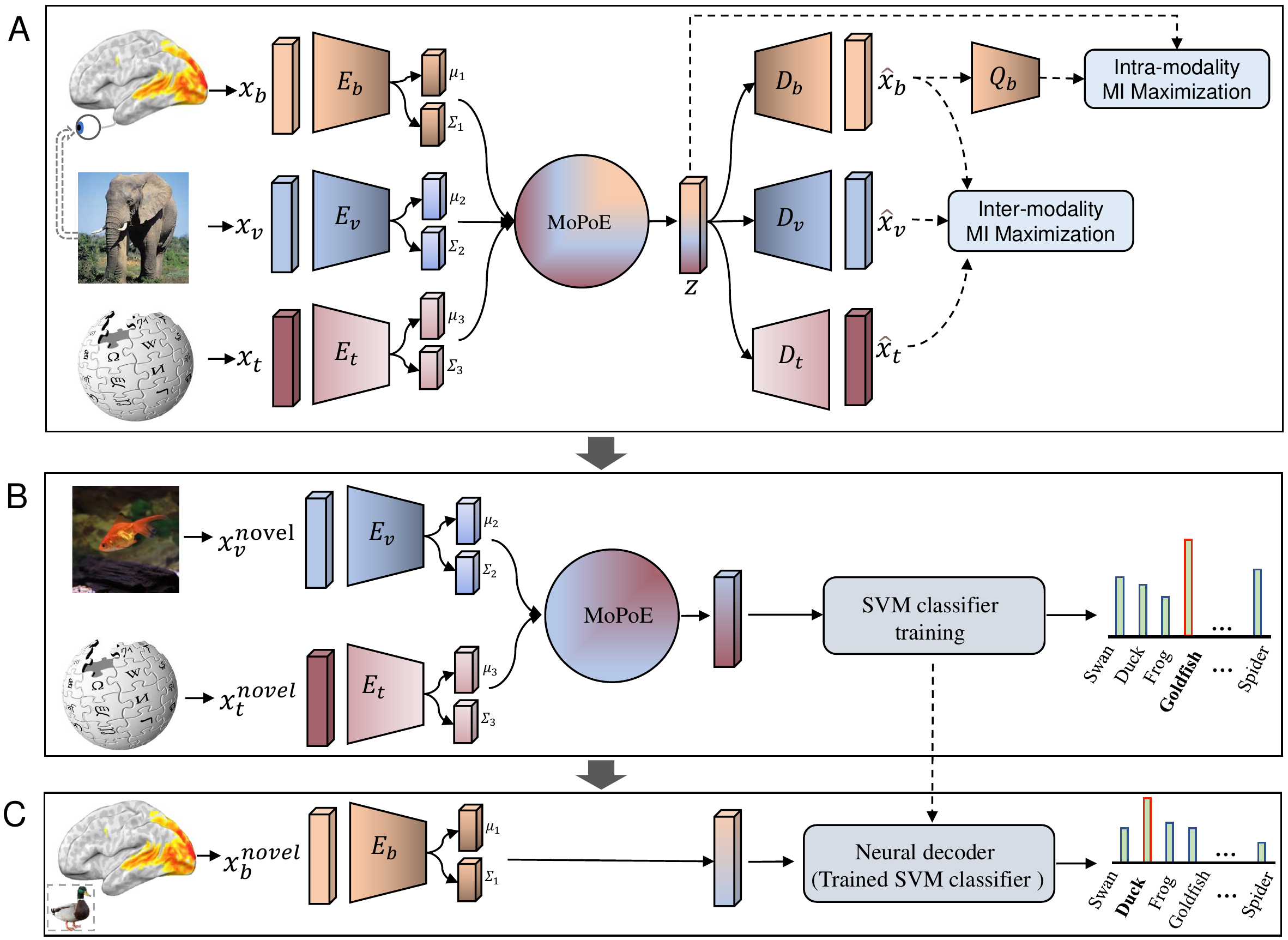}
		\vspace{-4pt}
		\caption{
			\textbf{(A). }Brain-Image-Text joint representation learning. The model works in two collaborative parts---multi-modality joint modeling and (both intra- and inter-modality) MI regularization. In this step, trimodal data from the \emph{seen} classes and the large scale bimodal (image-text pairs) or unimodal (image/text) data from the \emph{novel} classes are used for training. \textbf{(B).} Latent space classifier training. A SVM classifier is trained from the latent representations of visual and textual features of the novel classes. Note that the encoders $E_v$ and $E_t$ are frozen in this step, and only the SVM classifier (in grey module) will be optimized. \textbf{(C).} Decoding neural representations. The latent representations of the brain activity of the novel classes are passed to the neural decoder (i.e., the trained SVM classifier) to obtain the decoded visual categories.  In this step, the encoder $E_b$ and the SVM classifier are always frozen, hence the label predicting space is identical to that in (B). The reason why the SVM classifier can be generalized from (B) to (C) is that the latent representations of these three modalities have been aligned in (A) thought the MoPoE formulation.
		}
		\vspace{-4pt}
		\label{fig:framework}
	\end{figure*}
	
	\subsection{High-level overview of the proposed BraVL model}
	
	Fig. \ref{fig:framework}A shows the overall architecture of the proposed BraVL model. The model works in two collaborative parts---multi-modality joint modeling and MI regularization:
	\begin{itemize}
		\item \noindent \textbf{Multi-modality joint modeling.} Based on the \emph{Mixture-of-Products-of-Experts} (MoPoE) formulation \cite{sutter2021generalized}, we develop a multimodal auto-encoding variational Bayesian model that enables us to utilize the visual and textual features jointly to enhance the brain activity representation learning and downstream novel class neural decoding performance. Specifically, we use three modality-specific encoding networks $E_b, E_v$ and $E_t$ to transform the unimodal features $\x_b, \x_v$ and $\x_t$ into the joint latent representation $\z$, which is then passed through three modality-specific decoding networks $D_b, D_v$ and $D_t$ for feature reconstruction, respectively. 
		
		\item \noindent \textbf{Mutual information (MI) regularization.} The MI at two levels---intra-modality level and inter-modality level are maximized simultaneously. The former is approximated by its variational lower bound \cite{agakov2004algorithm} and the latter is achieved through introspective cross-modal contrastive learning. The MI at intra-modality level is used as a consistent regularizer to force the joint latent representation $\z$ to have a strong relationship with observations $\x_b, \x_v$ and $\x_t$, and hence learn useful joint representations. The MI at inter-modality level is used as a contrastive regularizer, which allows us to train the model not only by the commonality between modalities, but also by the distinction between ``related'' and ``unrelated'' multimodal data pairs, improving data efficiency in the case of limited brain activity data.	
	\end{itemize}

	\subsection{Multi-modality joint modeling}\label{joint_modeling}
	Variational auto-encoders (VAEs) \cite{kingma2013auto, rezende2014stochastic} are latent variable models for scalable unsupervised representation learning. For single modality data, they define the marginal log-likelihood via a
	latent variable $\z$: $\log p_{\theta}(\x)=\log \int p_{\theta}(\x|\z) p(\z) d \z$.
	Since exactly estimating log $p_{\theta}(\x)$ is typically intractable, the VAE instead maximizes a tractable Evidence Lower BOund (ELBO):
	\begin{align}
		\nonumber 	\log p_{\theta}(\x) &\geq  \mathbb{E}_{q_\phi(\z|\x)}[\log p_\theta(\x|\z)] - D_{KL}\left[q_\phi(\z|\x)\|p(\z)\right] \\
		& = \operatorname{ELBO}(\x),
	\end{align}
	where $q_\phi(\z|\x)$ is an approximate posterior distribution represented by the encoder network parametrized by $\phi$, and $p_\theta(\x|\z)$ is the decoder network parametrized by $\theta$.

	For the neural decoding task, we condition the VAE on the brain activity, visual features and corresponding textual features, forming a multimodal VAE. Formally, the joint marginal log-likelihood and its multimodal variational $\operatorname{ELBO}$  can be similarly given by:
	\begin{align} \label{ELBO_X}
		\nonumber	\log p_{\Theta}(\mathbb{X}) & \geq \mathbb{E}_{q_{\Phi}(\boldsymbol{z} \mid \mathbb{X})}\bigg[\sum_{\x_{m} \in \mathbb{X}} \log p_{\theta_m}\left(\x_{m} | \z\right)\bigg]\\
		\nonumber		&\hspace{3cm}-D_{KL}\left[q_{\Phi}(\z |\mathbb{X}) \| p(\z)\right]\\
		&=	\operatorname{ELBO}(\mathbb{X}) \triangleq \mathcal{L}_{M}(\Theta,\Phi),
	\end{align}
	where $p_{\theta_m}(\x_{m} | \z)$ are the decoders parameterized by $\Theta=\{\theta_m \}$, $m\in\{b, v, t\}$, respectively, $\mathbb{X}=\{ \x_b, \x_v , \x_t \}$ on the \emph{seen} class dataset $\mathcal{D}^{seen}$, and  $\mathbb{X}=\{\x_v , \x_t \}$ on the \emph{novel} class dataset $\mathcal{D}^{novel}$. To maximize $\log p_\Theta(\mathbb{X})$, one approximates
	the intractable model joint posterior $p_\Theta(\z|\mathbb{X})$ with a variational joint posterior $q_\Phi(\z|\mathbb{X})$, allowing us to optimize $\operatorname{ELBO}(\mathbb{X})$.
	Recently, there have been three dominant strains for constructing  $q_\Phi(\z|\mathbb{X})$, namely, the \emph{Product of Experts} (PoE) \cite{MVAE_wu_2018}, \emph{Mixture of Experts} (MoE) \cite{shi2019variational} and \emph{Mixture-of-Products-of-Experts} (MoPoE) \cite{sutter2021generalized}, as shown in Table \ref{Table:posterior}.  Here, we choose MoPoE in the neural decoding tasks due to its powerful joint latent representation learning and cross-modality generation ability, and (\ref{ELBO_X}) can be further written as: 
	\begin{small}
		\begin{align} \label{L_M}
			\nonumber	\mathcal{L}_{M}(\Theta,\Phi) & = \mathbb{E}_{q_{\Phi}(\boldsymbol{z} \mid \mathbb{X})}\bigg[\sum_{\x_{m} \in \mathbb{X}} \log p_{\theta_m}\left(\x_{m} | \z\right)\bigg]\\
			&\hspace{0.3cm}-D_{KL}\bigg[\frac{1}{|\mathcal{P}(\mathbb{X})|} \sum_{\mathbb{X}_s \in \mathcal{P}(\mathbb{X})} \prod_{\x_m \in \mathbb{X}_s} q_{\phi_m}\left(\z | \x_{m}\right) \| p(\z)\bigg],
		\end{align}
	\end{small}
	where $\mathbb{X}_s$ is a possible subset of $\mathbb{X}$, and $\mathcal{P}(\mathbb{X})$ is a new set made up of all possible subsets of $\mathbb{X}$.

	\begin{table*}[!htbp]
		\co{3.0pt}
		\ra{1.5}
		\centering
		\caption{Overview of the variational joint posterior forms in family of multimodal VAEs.}
		\vspace{-6pt}
		\resizebox{16.1cm}{!}{
			\begin{tabular}{p{2.0cm}|p{5.5cm}|p{2.9cm}|p{2.9cm}|p{2.8cm}}
				\thickhline
				Model                                      &Decomposition of $q_\Phi(\z|\mathbb{X})$  &Aggregate modalities  &Multimodal posterior   & Missing modalities                \\ \hline
				PoE \cite{MVAE_wu_2018}                    & $\prod_{\x_m \in \mathbb{X}} q_{\phi_m}\left(\z|\x_{m}\right)$                                     & \qquad\qquad \cmark                 & \qquad\qquad\xmark                   & \qquad\qquad(\cmark)             \\
				MoE  \cite{shi2019variational}             & $\frac{1}{|\mathbb{X}|} \sum_{\x_m \in \mathbb{X}} q_{\phi_m}\left(\z | \x_{m}\right)$           & \qquad\qquad\xmark                 & \qquad\qquad\cmark                   & \qquad\qquad\cmark             \\ 
				MoPoE  \cite{sutter2021generalized}        & $\frac{1}{|\mathcal{P}(\mathbb{X})|} \sum_{\mathbb{X}_s \in \mathcal{P}(\mathbb{X})} \prod_{\x_m \in \mathbb{X}_s} q_{\phi_m}\left(\z | \x_{m}\right)$                                      & \qquad\qquad \cmark                 & \qquad\qquad \cmark                  & \qquad\qquad \cmark           \\ 
				\thickhline
			\end{tabular}
		}
		\vspace{-6pt}
		\label{Table:posterior}
	\end{table*}

	Although maximizing $\operatorname{ELBO}$ is a theoretically elegant objective for VAEs, having a high $\operatorname{ELBO}$ does not necessarily mean that consistent and useful latent representations were learned \cite{alemi2018fixing}. The issue
	is especially challenging when a multimodal VAE is adopted. Specifically, in multimodal VAEs, the optimal $\mathcal{L}_{M}(\Theta,\Phi)$ forces the variational joint posterior to be close to the prior, regardless of how expressively
	$q_\Phi(\z|\mathbb{X})$ is parameterized. This typically causes well-known posterior collapse issues:
	1) the joint latent variable $\z$ is independent from the observed data $\mathbb{X}$; 
	2) the reconstruction of $\mathbb{X}$ cannot benefit from the encoders, meaning that the decoders can ignore the conditioning on $\z$ \cite{bowman2016generating,chen2016variational};
	3) there is weak coherence between joint generation over all modalities in multimodal scenarios.
	To alleviate the above issues, we explicitly introduce two mutual information regularization terms into the original multimodal $\operatorname{ELBO}$ objective, as explained in Sec. \ref{MI}.
	
	\subsection{Mutual information (MI) regularization} \label{MI}
	
	For two random variables, if the MI between them is high, they are highly predictable of each other. This hints that we can regularize the original multimodal $\operatorname{ELBO}(\mathbb{X})$ objective in \eqref{L_M} by 1) maximizing the intra-modality MI between the joint latent variable $\z$ and individual modalities and 2) maximizing the inter-modality MI across multiple modalities. Below, we will introduce them separately.
	
	\subsubsection{Intra-modality MI maximization}
	We first introduce intra-modality MI regularization that prefers high mutual information between the joint latent variable $\z$ and the individual modalities into the original $\operatorname{ELBO}$ objective. This encourages the model to make effective use of the joint latent variable and alleviate the posterior collapse issue. More formally, for $m\in\{b, v, t\}$, the MI between samples of the posterior distribution of joint latent variable $\z$ and observation $\x_m$ is defined as $I(\z; \x_m) = H(\z) - H(\z | \x_m)$.
	Here,  we derive an accurate and straightforward variational lower bound on MI as follows:
	\begin{align} \label{lower_bound}
		\nonumber	&I(\z ; \x_m)\\
		\nonumber	& = H(\z) - H(\z | \x_m) \\
		\nonumber	&= \mathbb{E}_{\x_m \sim p_{\theta_m}(\x_m | \z)}\left[\mathbb{E}_{\z \sim q_{\Phi}(\z|\mathbb{X})}[\log p(\z|\x_m)]\right]+H(\z) \\
		\nonumber	&= \mathbb{E}_{\x_m \sim p_{\theta_m}(\x_m \mid \z)}\big[\underbrace{D_{K L}(p(\cdot|\x_m) \| Q_{\psi_m}(\cdot|\x_m))}_{\geq 0}\\ 
		\nonumber	&\quad+\mathbb{E}_{\z \sim q_{\Phi}(\z|\mathbb{X})}\left[\log Q_{\psi_m}\left(\z| \x_m\right)\right]\big]+H(\z) \\ 
		&\geq \mathbb{E}_{\x_m \sim p_{\theta_m}(\x_m | \z)}\left[\mathbb{E}_{\z \sim q_{\Phi}(\z|\mathbb{X})}\left[\log  Q_{\psi_m}\left(\z | \x_m\right)\right]\right]+H(\z),
	\end{align}
	where $Q_{\psi_m}$ is some auxiliary distribution implemented by a deep neural network parameterized by $\psi_m$. As shown in Fig. \ref{fig:framework}A, $Q_{\psi_m}$ takes the output of $D_m$ as input. This lower bound is tight 
	when $Q_{\psi_m}(\z|\x_m)=p(\z|\x_m)$ which can be achieved by optimizing $\psi_m$.  Note that, we have omitted $Q_{v}$ and $Q_{t}$ in Fig. \ref{fig:framework}A for brevity.
	
	The problem of the above obtained variational lower bound is that it requires sampling $\z$ from the approximated joint posterior $q_{\Phi}(\z|\mathbb{X})$ in the inner expectation. Fortunately, we can overcome this by applying the lemma \ref{thelemma} proved in InfoGAN~\cite{chen2016infogan}, and rewrite (\ref{lower_bound}) as 
	\begin{align}
		\nonumber	I(\z ; \x_m)&\geq \mathbb{E}_{\z \sim q_{\Phi}(\z|\mathbb{X}),\ \x_m \sim p_{\theta_m}(\x_m | \z)}\left[\log Q_{\psi_m}\left(\z | \x_m\right)\right] \\
		& \quad +H(\z).
	\end{align}
	\begin{lemma}
		\label{thelemma}
		For random variables $X, Y$ and function $f(x, y)$ under suitable regularity conditions: 
		\begin{align} 
			\E_{x \sim X, y \sim  Y|x} [f(x, y)] = 	\E_{x \sim X, y \sim  Y|x, x' \sim  X|y} [f(x', y)].
		\end{align}
	\end{lemma}

	Finally, we define the intra-modality MI maximization regularizer $\mathcal{L}_{intra}$ for multimodal VAE as 
	\begin{align}\label{eq_intra}
		\nonumber&\mathcal{L}_{intra}(\Theta,\Phi,\Psi)= \\
		&\hspace{0.2cm} \sum_{m} \mathbb{E}_{\z \sim q_{\Phi}(\z|\mathbb{X}),\ \x_m \sim p_{\theta_m}(\x_m | \z)}\left[\log Q_{\psi_m}\left(\z | \x_m\right)\right] +3 H(\z),
	\end{align}
	where  $\Psi=\{\psi_m\}$, $m\in\{b, v, t\}$ on the \emph{seen} class dataset $\mathcal{D}^{seen}$, and  $m\in\{v, t\}$ on the \emph{novel} class dataset $\mathcal{D}^{novel}$.

	\subsubsection{Inter-modality MI maximization}
	The above intra-modality MI regularization focuses on mitigating the issues in learning consistent/useful joint representations in multimodal VAE. There is no guarantee that the joint generation over all modalities is strongly coherent due to modality-specific random noise and potential overfitting problems. Therefore, we further introduce inter-modality point-wise MI regularization $\mathcal{L}_{inter}$ that prefers high mutual information across multiple modalities. This allows us to suppress modality-specific noise that is irrelevant to the task, improve data efficiency in the case of limited brain activity, and hence keep generation coherence as much as possible.
	
	On the \emph{seen} class dataset $\mathcal{D}^{seen}$, $\mathcal{L}_{inter}$ is defined as:
	\begin{small}
		\begin{align} \label{L_inter}
			\nonumber	&\mathcal{L}_{inter}(\Theta,\Phi)\\
			\nonumber	&\hspace{0.3cm}=I\left(\x_{b}, \x_{v}; \x_{t}\right)+I\left(\x_{b} ; \x_{v}, \x_{t}\right)+I\left(\x_{b}, \x_{t} ; \x_{v}\right)\\
			\nonumber	&\hspace{0.3cm}=\log \frac{P_{\Theta}\left(\x_{b}, \x_{v}, \x_{t}\right)}{P_{\Theta}\left(\x_{b}, \x_{v}\right) p_{\Theta}\left(\x_{t}\right)} + \log \frac{P_{\Theta}\left(\x_{b}, \x_{v}, \x_{t}\right)}{P_{\Theta}\left(\x_{b}\right) p_{\Theta}\left(\x_{v}, \x_{t}\right)} \\
			\nonumber	& \hspace{0.6cm} + \log \frac{P_{\Theta}\left(\x_{b}, \x_{v}, \x_{t}\right)}{P_{\Theta}\left(\x_{b}, \x_{t}\right) p_{\Theta}\left(\x_{v}\right)}\\
			\nonumber	& \hspace{0.3cm}\approx\log \frac{P_{\Theta}\left(\x_{b}, \x_{v}, \x_{t}\right)}{ \sum\limits_{\x_{t}^{\prime}} P_{\Theta}\left(\x_{b}, \x_{v}, \x_{t}^{\prime}\right) \sum\limits_{\x_{b}^{\prime}} \sum\limits_{\x_{v}^{\prime}} P_{\Theta}\left(\x_{b}^{\prime}, \x_{v}^{\prime}, \x_{t}\right)} \\
			\nonumber	& \hspace{0.6cm} +\log \frac{P_{\Theta}\left(\x_{b}, \x_{v}, \x_{t}\right)}{\sum\limits_{\x_{v}^{\prime}} \sum\limits_{\x_{t}^{\prime}} P_{\Theta}\left(\x_{b}, \x_{v}^{\prime}, \x_{t}^{\prime}\right) \sum\limits_{\x_{b}^{\prime}} P_{\Theta}\left(\x_{b}^{\prime}, \x_{v}, \x_{t}\right)}\\
			\nonumber	& \hspace{0.6cm} +\log \frac{P_{\Theta}\left(\x_{b}, \x_{v}, \x_{t}\right)}{ \sum\limits_{\x_{v}^{\prime}} P_{\Theta}\left(\x_{b}, \x_{v}^{\prime}, \x_{t}\right) \sum\limits_{\x_{b}^{\prime}} \sum\limits_{\x_{t}^{\prime}} P_{\Theta}\left(\x_{b}^{\prime}, \x_{v}, \x_{t}^{\prime}\right)}\\
			\nonumber	&\hspace{0.3cm} = 3 \log P_{\Theta}\underbrace{\left(\x_{b}, \x_{v}, \x_{t}\right)}_{\text {\emph{{\color{black}positive}}}}  - \log \sum\limits_{\x_{t}^{\prime}} P_{\Theta}\underbrace{\left(\x_{b}, \x_{v}, \x_{t}^{\prime}\right)}_{\text {\emph{{\color{black}negative}}}} \\
			\nonumber	& \hspace{0.64cm} - \log \sum\limits_{\x_{b}^{\prime}} \sum\limits_{\x_{v}^{\prime}} P_{\Theta}\underbrace{\left(\x_{b}^{\prime}, \x_{v}^{\prime}, \x_{t}\right)}_{\text {\emph{{\color{black}negative}}}} - \log \sum\limits_{\x_{v}^{\prime}} \sum\limits_{\x_{t}^{\prime}} P_{\Theta}\underbrace{\left(\x_{b}, \x_{v}^{\prime}, \x_{t}^{\prime}\right)}_{\text {\emph{{\color{black}negative}}}} \\
			\nonumber	& \hspace{0.6cm} - \log \sum\limits_{\x_{b}^{\prime}} P_{\Theta}\underbrace{\left(\x_{b}^{\prime}, \x_{v}, \x_{t}\right)}_{\text {\emph{{\color{black}negative}}}}  - \log \sum\limits_{\x_{v}^{\prime}} P_{\Theta}\underbrace{\left(\x_{b}, \x_{v}^{\prime}, \x_{t}\right)}_{\text {\emph{{\color{black}negative}}}}\\
			& \hspace{0.6cm} - \log \sum\limits_{\x_{b}^{\prime}} \sum\limits_{\x_{t}^{\prime}} P_{\Theta}\underbrace{\left(\x_{b}^{\prime}, \x_{v}, \x_{t}^{\prime}\right)}_{\text {\emph{{\color{black}negative}}}},
		\end{align}
	\end{small}
	where $\x_{b}^{\prime} \in X_{b}^s\setminus\x_{b}$,  $\x_{v}^{\prime} \in X_{v}^s\setminus\x_{v}$ and $\x_{t}^{\prime} \in X_{t}^s\setminus\x_{t}$ are used to construct  \emph{negative} samples. We see in \eqref{L_inter} that maximizing $\mathcal{L}_{inter}$ can be decomposed to maximize the joint marginal likelihood of \emph{positive} samples and minimize the joint marginal likelihood of six kinds of \emph{negative} samples. This is equivalent to maximizing the difference between the joint marginal likelihoods of \emph{positive} and \emph{negative} samples. Interestingly, we also note that the first term of \eqref{L_inter} is identical to the original objective of multimodal VAE, and hence maximizing  $\mathcal{L}_{inter}$ introspectively forms contrastive learning in the model without introducing extra parameters.
	
	Since we do not have access to the exact joint marginal likelihood, we need to resort to some estimators to approximate the contrastive regularizer in \eqref{L_inter}. 
	For the first term of  \eqref{L_inter}, as in multimodal VAE, $\operatorname{ELBO}$ is a valid lower-bound estimator. Since the remaining terms are expected to be minimized in model training, we need to employ an upper-bound estimator. Here, we use the $\chi$-upper-bound ($\operatorname{CUBO}$) estimator proposed in \cite{cubo}.  The relationships between $\operatorname{ELBO}$, the (log) joint marginal likelihood $\log p_{\Theta}(\mathbb{X})$ and $\operatorname{CUBO}$ are as follows:
	\begin{small}
		\begin{align}
			\operatorname{ELBO} \leq \log p_{\Theta}(\mathbb{X}) \leq \underbrace{\mathbb{E}_{\left\{\z_{k}\right\}_{1}^{K} \sim q_{\Phi}}\left[\log \sqrt[2]{\frac{1}{K} \sum_{k=1}^{K}\left(\frac{p_{\Theta}\left(\boldsymbol{z}_{k}, \mathbb{X}\right)}{q_{\Phi}\left(\boldsymbol{z}_{k} | \mathbb{X}\right)}\right)^{2} } \right]}_{\text {CUBO}}
		\end{align}
	\end{small}
	
	On the \emph{novel} class dataset $\mathcal{D}^{novel}$, where only visual and textual features are provided, it is straightforward to obtain
	\begin{small}
		\begin{align}
			\nonumber	&\mathcal{L}_{inter}(\Theta,\Phi) =I\left( \x_{v}; \x_{t}\right)\\
			\nonumber	&=\log \frac{P_{\Theta}\left(\x_{v}, \x_{t}\right)}{P_{\Theta}\left(\x_{v}\right) p_{\Theta}\left(\x_{t}\right)}\\
			\nonumber	& \approx\log \frac{P_{\Theta}\left(\x_{v}, \x_{t}\right)}{\sum\limits_{\x_{t}^{\prime}} P_{\Theta}\left(\x_{v}, \x_{t}^{\prime}\right) \sum\limits_{\x_{v}^{\prime}} P_{\Theta}\left( \x_{v}^{\prime}, \x_{t}\right) }\\
			& =  \log P_{\Theta}\underbrace{\left(\x_{v}, \x_{t}\right)}_{\text {\emph{{\color{black}positive}}}}  - \log \sum\limits_{\x_{t}^{\prime}} P_{\Theta}\underbrace{\left(\x_{v}, \x_{t}^{\prime}\right)}_{\text {\emph{{\color{black}negative}}}}   - \log \sum\limits_{\x_{v}^{\prime}} P_{\Theta}\underbrace{\left( \x_{v}^{\prime}, \x_{t}\right)}_{\text {\emph{{\color{black}negative}}}} ,
		\end{align}
	\end{small}
	where $\x_{v}^{\prime} \in X_{v}^n\setminus\x_{v}$ and $\x_{t}^{\prime} \in X_{t}^n\setminus\x_{t}$ are used to construct \emph{negative} samples.

	\vspace{-2pt}
	\subsection{Overall objective and training}
	\textbf{Objective.} 
	We combine $\mathcal{L}_{M}$ with the two proposed MI regularizers $\mathcal{L}_{intra}$ and $\mathcal{L}_{inter}$ to form the objective
	\begin{align}\label{L_total}
		\nonumber	\max _{\Theta,\Phi} \max _{\Psi}\  & \mathcal{L}(\Theta, \Phi, \Psi) = \mathcal{L}_{M}(\Theta,\Phi) + \lambda_1 \mathcal{L}_{intra}(\Theta,\Phi,\Psi) \\
		& \hspace{3.5cm}  + \lambda_2\mathcal{L}_{inter}(\Theta,\Phi),
	\end{align}
	where $\lambda_1, \lambda_2$ are hyper-parameters that control the impact of MI maximization.
	For each training minibatch, the training process consists of two stages. In the first stage, we maximize the objective w.r.t. parameters $\Psi=\{\psi_b,\psi_v,\psi_t\}$ to make the variational lower bound on intra-modality MI tighter (i.e., minimize the gap between $Q_{\psi_m}(\z|\x_m)$ and $p(\z|\x_m)$). In the second stage,  we maximize the objective w.r.t. parameters $\Theta$ and $\Phi$ to train it using both intra- and inter-modality MI regularizers.	Below, we summarize the training procedure in Algorithm~\ref{alg:1}.
	
	\vspace{-1.5pt}
	\begin{algorithm}[ht!]
		\small
		\SetAlgoLined
		\KwIn{Datasets $\mathcal{D}^{seen}$, $\mathcal{D}^{novel}$, hyper-parameters $\lambda_1$, $\lambda_2$}
		\KwOut{$\Theta, \Phi, \Psi$}
		\Repeat{\itshape convergence of parameters $\Theta, \Phi, \Psi$}{
			Random minibatch drawn from $\mathcal{D}^{seen}$ or $\mathcal{D}^{novel}$\\
			\itshape $\bigcdot$ Stage 1: Tightening lower bound of intra-modality MI\\
			\upshape
			\ \ \	Maximize $\mathcal{L}_{intra}(\Theta, \Phi, \Psi)$ w.r.t. $\Psi$\\
			\itshape $\bigcdot$ Stage 2: Joint Training with MI-maximization\\ 
			\upshape
			\ \ \	Maximize $\mathcal{L}(\Theta, \Phi, \Psi)$ w.r.t. $\Theta$ and $\Phi$
		}
		\Return $\Theta, \Phi, \Psi$
		\caption{\small Training BraVL}
		\label{alg:1}
	\end{algorithm}
	\vspace{-1.0pt}
	
	\textbf{Semi-supervised learning by utilizing  extra data.} 
	In addition to the complete trimodal data (brain-image-text), we are also allowed to incorporate large-scale bimodal (image-text pairs) or unimodal (image/text) data obtained from extra databases. This is beneficial for the neural decoding field where brain activity data are scarce but visual image and textual data are plentiful.

	\textbf{SVM classifier training for novel class neural decoding.} 
	After training the BraVL model,  an additional classifier is trained from the latent vectors, as shown in Fig. \ref{fig:framework}B. Specifically, a Support Vector Machine (SVM) classifier is trained from
	the latent vectors, obtained from the visual and textual features of the novel classes, to the class labels of these novel classes. At test time, the latent vectors obtained from the brain activity are passed
	to the trained SVM classifier (neural decoder), as shown in Fig. \ref{fig:framework}C.

	\vspace{-1.0pt}
	\section{Experiments} \label{experiments}
	
	\begin{table*}[!htbp]
		\small
		\ra{1.2}
		\co{2.3pt}
		\centering
		\caption{Characteristics of the brain-visual-linguistic datasets used in our experiments. For brain features on the GOD-Wiki and DIR-Wiki, distinct subjects have different data dimensions.}
		\vspace{-6pt}
		\resizebox{17.5cm}{!}{
			\begin{tabular}{c | c c c c c c c c}
				\thickhline
				Dataset               & Training samples  & Test samples     & Brain features                      & Visual features    & Textual features   & Subjects      &Seen classes     &Novel classes          \\ \thickhline
				GOD-Wiki              & 1200              & 1750             & 726/945/1034/998/1009               & 4996               & 2560               & 5             &150              &50       \\
				DIR-Wiki              & 6000              & 1200             & 1401/1616/1355                      & 4996               & 2560               & 3             &150              &50    \\ 
				ThingsEEG-Text        & 16540             & 16000            & 561                                 & 100                & 512                & 10            &1654             &200    \\
				\thickhline
			\end{tabular}
		}
		\vspace{-6pt}
		\label{Table:datasets}
	\end{table*}
	\begin{figure*}[t!]
		\centering 
		\includegraphics[scale=0.7]{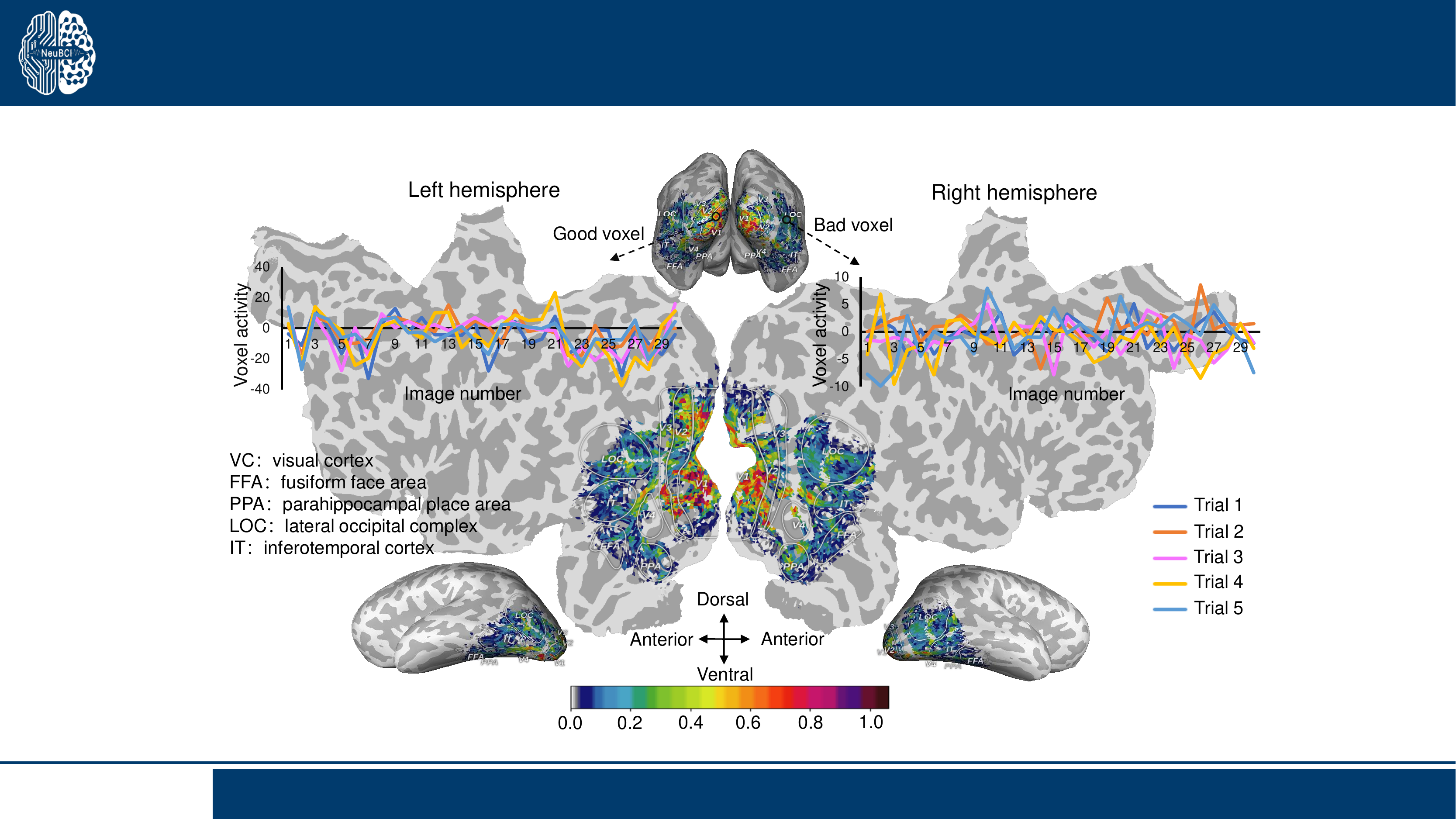}
		\vspace{-8pt}
		\caption{The voxel stability maps in the visual cortex. For each voxel in each brain Regions of Interest (ROI), the stability score is quantified as the mean Pearson correlation coefficient across all pairwise combinations of the five repetitions (trials). Then, the stability score for each voxel is projected onto a cortical flat map constructed for an example subject (Subject 3 of DIR-Wiki dataset) using Pycortex \cite{gao2015pycortex}. For each ROI, only the top 15\% voxels with larger stability scores are used in neural decoding tasks. 
		}
		\vspace{-5pt}
		\label{fig:Stability}
	\end{figure*}
	
	\subsection{Brain-Visual-Linguistic datasets.}
	We refer to our newly constructed trimodal datasets as GOD-Wiki, DIR-Wiki and ThingsEEG-Text, which are based on the public GOD~\cite{horikawa2017generic}, DIR \cite{shen2019deep} and ThingsEEG \cite{gifford2022large} datasets, respectively. The characteristics of the datasets used in our experiments are summarized in Table \ref{Table:datasets}.
	
	\textbf{GOD-Wiki:}\ \  
	GOD dataset~\cite{horikawa2017generic}\footnote{https://figshare.com/articles/dataset/Generic\_Object\_Decoding/7387130} provides the brain fMRI recordings of five subjects when they are presented with ImageNet images. 
	The GOD comprises 1250 distinct images drawn from 200 selected ImageNet categories. Among them, 150 categories provide 1200 training images (eight from each category), and the remaining 50 categories provide 50 test images (one from each category). 
	The training and test image stimuli were repeated 1 and 35 times, respectively, leading to 1200 and 1750 brain activity instances, respectively. The fMRI data were collected using a 3.0-Tesla Siemens MAGNETOM Trio scanner with repetition time (TR) = 3 s, voxel size, 3 $\times$ 3 $\times$ 3 mm$^{3}$. The acquired fMRI data underwent three-dimensional motion
	correction using SPM5 (http://www.fil.ion.ucl.ac.uk/spm). The data were then coregistered to the within-session high-resolution anatomical image of the
	same slices used for echo-planar imaging (EPI) and subsequently to the whole-head high-resolution anatomical image. Visual areas V1, V2, V3, and V4 were delineated following the standard retinotopy experiment.
	The lateral occipital complex (LOC), fusiform face area (FFA), and parahippocampal place area (PPA) were identified using conventional functional localizers. More details about the fMRI data preprocessing can be found in the original paper~\cite{horikawa2017generic}. Here, we considered 4648 voxels from the brain visual areas V1, V2, V3, V4, LOC, FFA and PPA, provided by~\cite{horikawa2017generic}.
	
	To obtain the corresponding text descriptions for the visual categories in GOD-Wiki, we adopt a semi-automatic method for corresponding Wikipedia article extraction. Specifically, we first create an automatic matching of ImageNet classes to their corresponding Wikipedia pages.\footnote{https://dumps.wikimedia.org/backup-index.html (enwiki-20200120).} Similar to \cite{bujwid-sullivan-2021-large}, the matching is based on the similarity between the synset words of ImageNet classes and Wikipedia titles, as well as their ancestor categories. Unfortunately, such matching occasionally produces false-positives since classes with similar names might represent very different concepts. To ensure high-quality matching between visual and linguistic features, we then manually removed the mismatched articles when constructing the trimodal dataset. 
	
	\textbf{DIR-Wiki:}\ \  
	DIR dataset \cite{shen2019deep}\footnote{https://openneuro.org/datasets/ds001506/versions/1.3.1} contains 1250 images that are identical to those used in the GOD dataset. Three healthy subjects were involved in the image presentation experiment. Since the training and test image stimuli were repeated 5 and 24 times respectively, the training set of the DIR dataset consists of a larger number of image-fMRI pairs (5 $\times$ 1200 samples) compared to the GOD dataset. The test set contains 1200 (24 $\times$ 50) brain activity instances. The fMRI data were collected using a 3.0-Tesla Siemens MAGNETOM Verio scanner. An interleaved T2$^{\star}$-weighted gradient-EPI scan was performed to acquire functional images covering the
	entire brain (TR = 2 s; voxel size, 2 $\times$ 2 $\times$ 2 mm$^{3}$; number of slices, 76). Similar to the GOD dataset, the fMRI data were then subjected to three-dimensional motion correction, coregistered to the high-resolution anatomical images and regions of interests (ROIs) selection. More fMRI data preprocessing information can be found in \cite{shen2019deep}.
	We used the fMRI voxels from the brain visual areas V1, V2, V3, V4, LOC, FFA, PPA and IT. Since DIR-Wiki and GOD-Wiki share the identical set of visual stimuli, we use exactly the same Wikipedia text descriptions mentioned above for DIR-Wiki. 
	
\begin{figure}[!htbp]
		\centering 
		\vspace{-5pt}
		\includegraphics[scale=0.44]{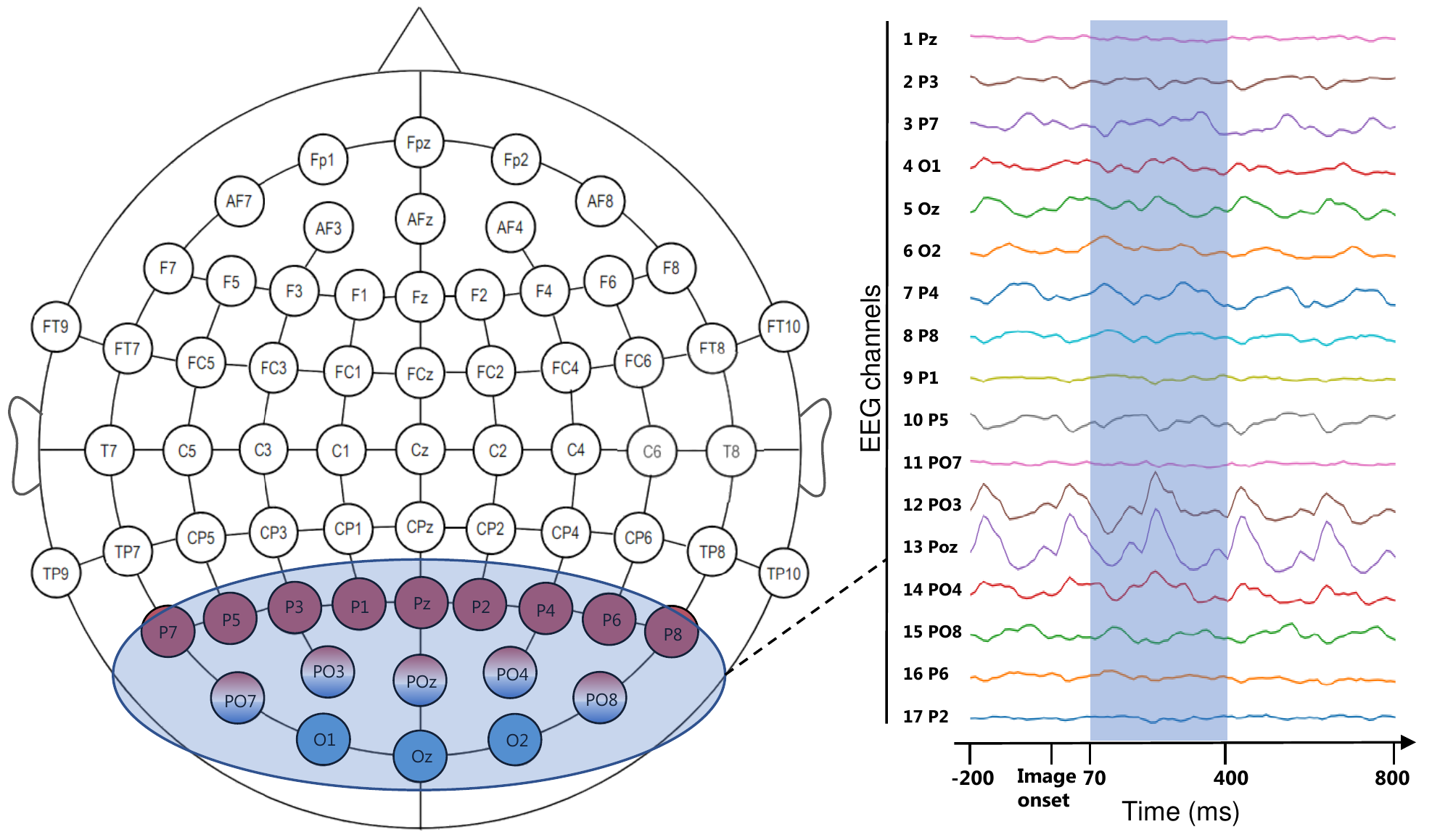}
		\vspace{-7pt}
		\caption{The EEG channels and time window used in our experiments.}
		\vspace{-2pt}
		\label{fig:eeg_channels}
	\end{figure}

	\textbf{ThingsEEG-Text:}\ \  
	ThingsEEG \cite{gifford2022large}\footnote{https://osf.io/3jk45/} is a large millisecond resolution EEG dataset collected by using a time-efficient rapid serial visual presentation (RSVP) paradigm. It contains 10 human subjects' EEG responses to 16,740 natural image stimuli taken from the THINGS database \cite{hebart2019things}. These 16740 images are composed of 1654 training classes (10 images per class) and 200 test classes (1 image per class). Each training image was presented 4 times, and each test image was presented 80 times, for a total of 82,160 image trials per subject. The EEG data were collected using a 64-channel EASYCAP equipment. As shown in Fig. \ref{fig:eeg_channels}, we used the preprocessed EEG data (time series from 70ms to 400ms with respect
	to image onset, down-sampled from 1000Hz to 100Hz, selected 17 channels overlying occipital and parietal cortex\footnote{O1, Oz, O2, PO7, PO3, POz, PO4, PO8, P7, P5, P3, P1, Pz, P2, P4, P6, P8}) provided by the authors \cite{gifford2022large}. Since the ThingsEEG dataset already provided 1000-dimensional visual features extracted by different pre-trained models (e.g., CORnet-S \cite{kubilius2019brain}), we used the top 100 principal components of the extracted CORnet-S features rather than extracting them again. For textual features, we first generated the corresponding textual description for each image by using a state-of-the-art image captioning model \cite{li2022blip} and then extracted their textual embeddings with CLIP \cite{radford2021learning}.
	
	\textbf{Extra visual-textual pairs:}\ \  As the extra data, we collected Wikipedia articles for 1000 categories selected from WordNet 3.0 and randomly selected 10, 20 or 50 images from the ImageNet dataset for each category, thus obtaining a total of 10k, 20k or 50k extra visual-textual data pairs. Unless otherwise stated, the results on the GOD-Wiki and DIR-Wiki (reported in Sec. \ref{results}) have been powered by the 10k extra data.

	\subsection{Implementation detail.}
	As shown in Table \ref{Table:BraVL_Architecture}, all encoders and decoders in BraVL are multi-layer perceptrons (MLPs) with two hidden layers (partly implemented by MindSpore). 
	On the GOD-Wiki and DIR-Wiki datasets, each hidden layer has 512 units for the brain modality,  2048 units for the visual modality and 512 units for the textual modality. The latent dimension is 32. The model is trained for 100 epochs by using the Adam optimizer \cite{kingma2014adam} with a learning rate of 1e-4 and a batch size of 512 on both fMRI-based datasets. 
	For seen classes, each training batch consists of (\emph{brain activity}, \emph{visual features}, \emph{textual features}) tuples, while for novel classes, each training batch consists of (\emph{visual features}, \emph{textual features}) pairs. 
	For the KL-divergence in the multimodal $\operatorname{ELBO}$,  we use an annealing scheme \cite{MVAE_wu_2018}, in which we increase its weight by a rate of 0.01 per epoch until it is equal to 1. 
	On DIR-Wiki, as shown in Fig. \ref{fig:Stability}, the voxel stability selection rate is 15\%, whereas on GOD-Wiki, we do not use voxel stability selection since there is only one trial for each stimulus in the training set. After performing 5-fold cross-validation with a grid search strategy on the training data (using \emph{sklearn} tools), we implemented the SVM classifier with RBF kernel parameter `$gamma$'=1e-5 and default setting of other parameters. The same SVM parameter settings were used for all subjects. The hyper-parameters $\lambda_1=0.001$, $\lambda_2=0.001$, $K=30$ for the $\operatorname{CUBO}$ estimators, and we empirically tuned them on one subject and then applied them to all other subjects.
	\begin{table}[!htbp] \centering
		\ra{1.0}
		\co{4.0pt}
		\vspace{-4pt}
		\caption{ Detailed BraVL architecture. On DIR-Wiki, $d_b$  is equal to 1401, 1616 and 1355 for Subjects 1, 2 and 3, respectively. On GOD-Wiki, $d_b$  is equal to 726, 945, 1034, 998 and 1009 for Subjects 1, 2, 3, 4 and 5, respectively.}
		\vspace{-8pt}
		\resizebox{8.2cm}{!}{
			\begin{tabular}{c|c|c}
				\toprule
				Brain Encoder $E_b$        & Visual Encoder $E_v$   & Textual Encoder $E_t$ \\
				\midrule
				Input:  $1\times d_b$ $\x_b$         & Input:  $1\times4996$ $\x_v$    & Input:  $1\times2560$ $\x_t$ \\
				MLP 512 units                   & MLP 2048 units              & MLP 512 units \\
				ReLU                  &  ReLU             &  ReLU\\
				MLP 512 units                   & MLP 2048 units             & MLP 512 units \\
				ReLU                  &  ReLU             & ReLU\\
				MLP output 32$\times$2              & MLP output 32$\times$2          & MLP output 32$\times$2\\
				\toprule
				Brain Decoder $D_b$        & Visual Decoder $D_v$   & Textual Decoder $D_t$ \\
				\midrule
				Latent  $1\times32$ $\bm{z}$        & Latent  $1\times32$ $\bm{z}$     & Latent  $1\times32$ $\bm{z}$\\
				MLP 512 units                  & MLP 2048 units             & MLP 512 units \\
				ReLU                  &  ReLU             &  ReLU\\
				MLP 512 units                   & MLP 2048 units            & MLP 512 units \\
				ReLU                  &  ReLU             &  ReLU\\
				MLP $d_b$ units                     & MLP 4996 units                  & MLP 2560 units\\
				\bottomrule
			\end{tabular}
		}
		\vspace{-8pt}
		\label{Table:BraVL_Architecture}
	\end{table}

	\subsection{Results} \label{results}
	
	\subsubsection{Does language influence vision? }  
	
	\begin{table*}[!htbp]     \centering
		\ra{0.9}
		\co{3.0pt}
		\caption{Neural decoding accuracy (\%) of several methods that are trained with the visual (V), textual (T) and combined (V\&T) features, respectively. * denotes the BraVL's performance is significantly better than the compared method (paired t-test, p$<$0.05).}
		\vspace{-7pt}
		\resizebox{18.2cm}{!}{
			\begin{tabular}{@{}lc*{20}{l}@{}}
				\toprule
				& & \multicolumn{6}{l}{\makecell{DIR-Wiki}} & \multicolumn{10}{l}{\makecell{GOD-Wiki}} & \multicolumn{2}{l}{} \\
				\cmidrule(l){3-8} \cmidrule(l){9-18} 
				& & \multicolumn{2}{l}{\makecell{Subject 1}} & \multicolumn{2}{l}{\makecell{Subject 2}} & \multicolumn{2}{l}{\makecell{Subject 3}} & \multicolumn{2}{l}{\makecell{Subject 1}} & \multicolumn{2}{l}{\makecell{Subject 2}}  & \multicolumn{2}{l}{\makecell{Subject 3}} & \multicolumn{2}{l}{\makecell{Subject 4}} & \multicolumn{2}{l}{\makecell{Subject 5}} & \multicolumn{2}{l}{\makecell{\textbf{Average}}} \\
				\cmidrule(l){3-4} \cmidrule(l){5-6} \cmidrule(l){7-8}  \cmidrule(l){9-10}  \cmidrule(l){11-12}  \cmidrule(l){13-14} \cmidrule(l){15-16}   \cmidrule(l){17-18}  \cmidrule(l){19-20}   
				\textbf{Model}                            & \textbf{Modality}  & top-1          & top-5          & top-1          & top-5     & top-1 & top-5  & top-1 & top-5   & top-1 & top-5  & top-1 & top-5  & top-1 & top-5    & top-1& top-5 & top-1& top-5 \\
				\midrule
				CADA-VAE\cite{cada}                       & V                  & 27.34          & 54.35          & 16.33          & 47.85     & 20.19 & 52.78  & 5.66  & 28.25   & 6.01  & 35.31   & 16.51 & 48.76  & 9.17 & 35.39    & 6.01 & 28.77  &13.40$_{*}$  & 41.43$_{*}$          \\
				CADA-VAE\cite{cada}                       & T                  & 3.26           & 12.02          & 3.03           & 12.63     & 3.25  & 13.54  & 3.03  & 13.20   & 3.15  & 13.57   & 3.23 & 13.18   & 3.14 & 12.36    & 3.15 & 12.13  &3.16$_{*}$  & 12.83$_{*}$          \\
				CADA-VAE\cite{cada}                       & V\&T               & 30.16          & 64.54          & 29.29          & 55.45     & 23.82 & 60.84  & 6.31  & 35.70   & 6.45  & 40.12   & 17.74 & 54.34  & 12.17 & 36.64   & 7.45 & 35.04   &16.67$_{*}$  &47.83$_{*}$         \\
				\midrule
				MVAE\cite{MVAE_wu_2018}                   & V                  & 25.56          & 53.83          & 18.56          & 46.57     & 21.74 & 55.15  & 5.30  & 24.64   & 5.21  & 32.58   & 14.13 & 50.11  & 8.03 & 34.46   & 5.44 & 27.45     &13.0$_{*}$  &40.60$_{*}$        \\
				MVAE\cite{MVAE_wu_2018}                   & T                  & 3.22           & 11.92          & 3.50           & 13.12     & 3.46  & 14.01  & 3.30  & 13.25   & 3.08  & 13.55   & 3.31 & 13.46   & 3.35 & 12.08    & 2.87 & 12.36  &3.26$_{*}$  &12.97$_{*}$           \\
				MVAE\cite{MVAE_wu_2018}                   & V\&T               & 29.34          & 65.02          & 28.76          & 57.23     & 26.58 & 62.34  & 5.77  & 31.51   & 5.40  & 38.46   & 17.11 & 52.46  & 14.02 & 40.90   & 7.89 & 34.63   &16.86$_{*}$  &47.82$_{*}$                    
				\\ \midrule
				MMVAE\cite{shi2019variational}            & V                  & 30.43          & 61.45          & 20.54          & 50.11     & 30.35 & 65.72  & 5.41  & 23.38   & 5.39  & 34.15   & 13.76 & 33.55  & 10.62 & 37.53  & 5.02 & 30.60   &15.19$_{*}$  &42.06$_{*}$       \\
				MMVAE\cite{shi2019variational}            & T                  & 3.43           & 15.32          & 3.58           & 13.44     & 3.33  & 12.66  & 3.07  & 13.32   & 3.24  & 14.05   & 3.14 & 13.34   & 3.58 & 12.56    & 3.24 & 12.02   &3.33$_{*}$  &13.34$_{*}$    \\
				MMVAE\cite{shi2019variational}            & V\&T               & 33.0           & 65.0           & 29.41          & 59.33     & 30.66 & 67.0   & 6.63  & 38.74   & 6.60  & 41.03   & 22.11 & 56.28  & 14.54 & 42.45  & 8.53 & 38.14   &18.94$_{*}$  &51.0$_{*}$    
				\\ \midrule
				MoPoE-VAE\cite{sutter2021generalized}     & V                  & 29.41          & 63.91          & 22.25          & 52.08     & 29.5 & 64.58   & 5.20  & 26.41   & 7.42  & 31.37   & 14.05 & 39.77  & 9.25 & 35.42  & 6.37 & 32.62   &15.43$_{*}$  &43.27$_{*}$         \\
				MoPoE-VAE\cite{sutter2021generalized}     & T                  & 4.48           & 15.25          & 4.50           & 16.58     & 4.25 & 14.91   & 3.25  & 13.22   & 3.26  & 14.02   & 3.54 & 14.45   & 3.70 & 13.45    & 3.08 & 12.90 &3.76$_{*}$  &14.35$_{*}$ \\
				MoPoE-VAE\cite{sutter2021generalized}     & V\&T               & 36.27          & 70.95          & 33.66          & 68.10     & 33.25 & 71.66  & 8.45  & 44.05   &8.34   & 48.11   & 22.68 & 61.82  & 14.57 & 58.51   & 10.45& 46.40 		&20.96$_{*}$  &58.70$_{*}$ 	           
				\\ \midrule
				BraVL (ours)                              & V                  & 34.91          & 69.50          & 30.08          & 63.83     & 34.83 & 74.33  & 8.91  & 37.37   &8.51   & 42.68   & 18.17 & 60.85  & 14.20 & 53.08   & 11.02&42.45 &20.08$_{*}$  &55.51$_{*}$         \\
				BraVL (ours)                              & T                  & 4.50           & 16.64          & 4.75           & 17.21     & 4.66 & 15.48   & 3.57  & 13.65   & 3.74  & 14.88   & 3.26 & 13.85  & 3.88  & 13.74   & 3.31  & 13.91 &3.96$_{*}$  &14.92$_{*}$ \\
				BraVL (ours)                              & V\&T               & {\bftab 38.67} & {\bftab 72.50} & {\bftab 34.50} & {\bftab 68.33}  & {\bftab 36.33} & {\bftab 76.16}   & {\bftab 9.11} & {\bftab 46.80}  & {\bftab 8.91} & {\bftab 48.86}  & {\bftab 24.0} & {\bftab 62.06}   & {\bftab 15.08} & {\bftab 60.0}  & {\bftab 12.86} & {\bftab 47.94} & {\bftab 22.43} & {\bftab 60.33}  \\
				\bottomrule
			\end{tabular}
		}
		\label{tbl:comparision}
	\end{table*}
	\begin{figure*}[!htbp]
		\vspace{-2pt}
		\centering 
		\includegraphics[scale=0.61]{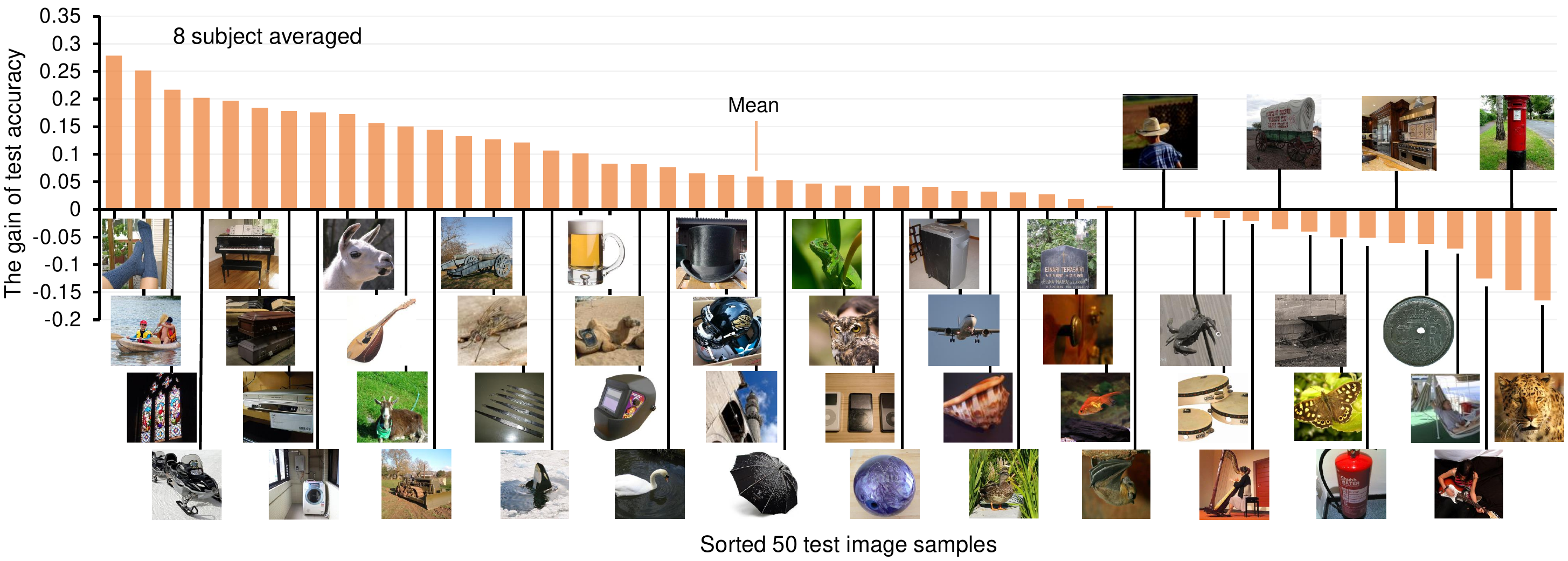}
		\vspace{-6pt}
		\caption{Qualitative comparison of the (top-1) decoding accuracy gain for each test class when the textual feature is added (produced by the BraVL method, averaged across all 8 subjects, sorted in descending order).
		}
		\vspace{-6pt}
		\label{fig:improvement}
	\end{figure*}
	Table \ref{tbl:comparision} shows the average neural decoding accuracy of several representative methods on the GOD-Wiki and DIR-Wiki datasets. The experimental results lead us to two significant observations. First,  models using the combination of visual and textual features (V\&T) perform much better than those using either of them alone. Remarkably, BraVL based on V\&T features leads to significant improvement in average top-5 accuracy on both datasets. 
	These results strongly suggest that although the stimuli presented to the subjects contained only visual information, it is conceivable that the subjects would subconsciously invoke the appropriate linguistic representations, which in turn influence visual processing.
	Thus, it is not surprising that multimodal models that learn joint representations of visual and linguistic features show superior decoding performance on `pure' visual neural response data.
	Second, BraVL trained with V\&T features achieves an average top-5 classification accuracy of 60.33\% and clearly outperforms the chance level (10\%) and well known recent models, such as CADA-VAE\cite{cada}, MVAE\cite{MVAE_wu_2018}, MMVAE\cite{shi2019variational}, and MoPoE-VAE\cite{sutter2021generalized}.
	This indicates that the fMRI signals triggered by natural visual stimuli
	encode enough information to significantly distinguish their visual categories under our specially designed multimodal decoding framework. 
	
	To enhance the understanding of textual feature influence for each test class, in Fig. \ref{fig:improvement}, we qualitatively compare the decoding accuracy gain for each test class when the textual feature is incorporated. As we can see, for most test classes, the consideration of textual features has positive effects, and the average Top-1 decoding accuracy improves by approximately 6\%.
	
	\begin{figure}[!htbp]
		\centering 
		\includegraphics[scale=0.61]{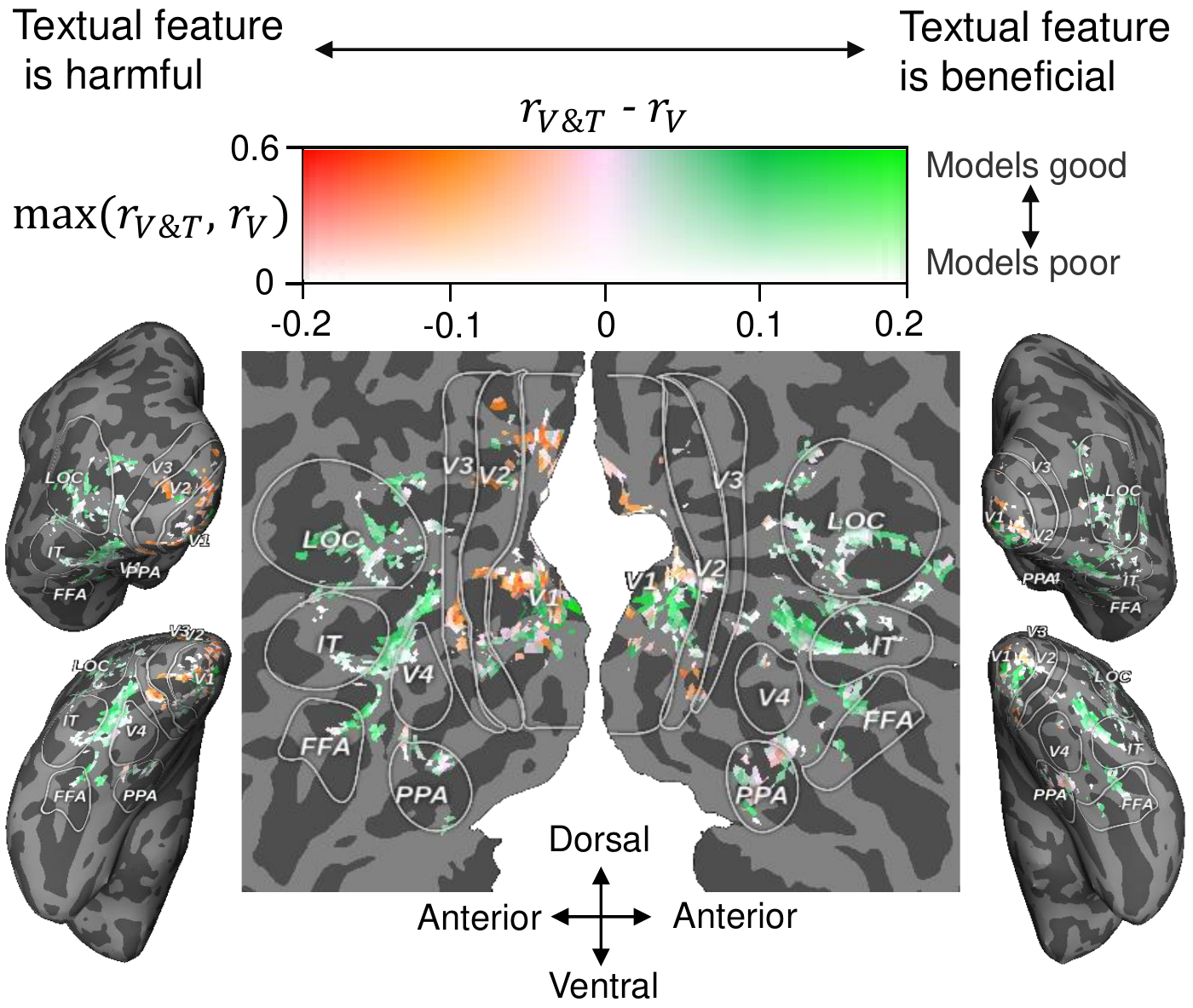}
		\vspace{-10pt}
		\caption{Projection of the textual feature contributions onto the visual cortex (produced by the BraVL method, Subject 3, DIR-Wiki). The green regions indicate that text features have a positive effect, and the red or orange regions indicate the opposite.}
		\vspace{-5pt}
		\label{fig:text_role_encoding}
	\end{figure}
	
	In addition to the neural decoding analysis, we also analyzed the contribution of textual features in terms of voxel-wise neural encoding (predicting corresponding brain voxel activity based on the visual or textual features). The voxel-wise encoding accuracy (we use $r$, the Pearson correlation coefficient between predicted brain activity and the ground-truth, as a metric) is visualized in Fig. \ref{fig:text_role_encoding}. It can be seen that for most high-level visual cortex (HVC, such as FFA, LOC and IT), incorporating textual features on top of visual features can improve the prediction accuracy of brain activity, while for most low-level visual cortex (LVC, such as V1, V2 and V3), incorporating textual features is not beneficial or even harmful. 
	From the perspective of cognitive neuroscience, our result is reasonable since it is generally believed that the HVC is responsible for processing the category information, movement information and other higher-level semantic information of objects, while the LVC is responsible for processing the orientation, contour and other low-level information \cite{dicarlo2012does}. Furthermore, a recent neuroscience study found that visual and linguistic semantic representations are aligned at the border of human visual cortex (i.e., `semantic alignment hypothesis') and that language-derived representations are located anterior to vision-derived representations \cite{popham2021visual}. Our results in Fig. \ref{fig:text_role_encoding} also support this hypothesis.

	\subsubsection{Are Wiki articles more effective than class names?}
	In Table \ref{tbl:name_article}, we compare different ways to incorporate textual features into our BraVL model for neural decoding tasks. We observe that regardless of which feature extraction method we use, using Wiki articles as class descriptions produces higher performance than using class names. In particular, using the ALBERT (xxlarge version) feature~\cite{lan2019albert} of Wiki articles in BraVL leads to an almost 12\% improvement (45.47\% vs. 57.04\%) in the top-5 accuracy on GOD-Wiki, which is far higher than that achieved in DIR-Wiki. We speculate that this improvement is influenced by the number of paired samples in the training set. These results indicate that Wikipedia class descriptions contain richer linguistic semantic features, and hence are a better choice for improving neural decoding accuracy than the class names most commonly used in prior works \cite{nishida2018decoding,vodrahalli2018mapping}.
	\begin{table}[!htbp]     \centering
		\ra{0.9}
		\co{4.0pt}
		\vspace{-6pt}
		\caption{Compare  Wiki articles and class names as textual inputs. The results are averaged across subjects. * denotes Wiki articles are significantly better than class names (paired t-test, p$<$0.05).}
		\vspace{-6pt}
		{\small
			\begin{tabular}{@{}ll*{6}{l}@{}}
				\toprule
				& & \multicolumn{2}{c}{\makecell{DIR-Wiki}} & \multicolumn{2}{c}{\makecell{GOD-Wiki}} \\
				\cmidrule(l){3-4} \cmidrule(l){5-6} 
				\textbf{Text data}            & \textbf{Features}             & top-1             & top-5                    & top-1                     & top-5 \\
				\midrule
				Class names                   & ALBERT~\cite{lan2019albert}   & 31.80             & 70.58                    & 11.49                     & 45.47             \\
				Wiki articles                 & ALBERT~\cite{lan2019albert}   & 36.11$_{*}$             & {\bftab 74.69}$_{*}$           & 13.09$_{*}$                     & {\bftab57.04}$_{*}$              \\
				Class names                   & GPT-Neo \cite{gpt-neo}        & 31.41             & 69.67                    & 10.77                     & 36.69              \\
				Wiki articles                 & GPT-Neo \cite{gpt-neo}        & {\bftab 36.50}$_{*}$    & 72.39$_{*}$                    &  {\bftab 13.39} $_{*}$          & 53.54$_{*}$              \\
				\bottomrule
			\end{tabular}
			\vspace{-3pt}
		}
		\label{tbl:name_article}
	\end{table}

	\subsubsection{Ablation study.} We conduct the first ablation experiments to discuss the role of MI regularization terms $\mathcal{L}_{intra}$ and  $\mathcal{L}_{inter}$, and the results are shown in Table \ref{tbl:Ablation}. 
	First, we present the results obtained using the complete losses. Then, on the basis of the complete method, we separately eliminate the intra-modality MI term (w/o $\mathcal{L}_{intra}$) or the inter-modality MI term (w/o $\mathcal{L}_{inter}$) from the total loss. Finally, we eliminate both the $\mathcal{L}_{intra}$ and $\mathcal{L}_{inter}$ terms from the objective of BraVL, and only use the multimodal $\operatorname{ELBO}$ loss (which is equivalent to MoPoE-VAE \cite{sutter2021generalized}) for training. We note the manifest performance degradation after removing either  $\mathcal{L}_{intra}$ or  $\mathcal{L}_{inter}$, and the results are even worse when removing two MI terms simultaneously, which shows the efficacy of our multimodal MI maximization framework.  We spot a clearer drop on GOD-Wiki than on DIR-Wiki, which implies that the inter-modality MI regularization term $\mathcal{L}_{inter}$ built on cross-modality contrastive learning leads to better data efficiency, especially when the number of training samples is small (GOD-Wiki contains only 1200 trimodal training samples, while DIR-Wiki contains 6000).

	\begin{table}[!htbp]     \centering
		\ra{0.9}
		\co{3.1pt}
		\vspace{-6pt}
		\caption{Ablation study of MI regularizers. * indicates that completed BraVL is significantly better than the compared one (paired t-test, p$<$0.05).}
		\vspace{-6pt}
		{\small
			\begin{tabular}{@{}ll*{6}{l}@{}}
				\toprule
				& & \multicolumn{2}{c}{\makecell{DIR-Wiki}} & \multicolumn{2}{c}{\makecell{GOD-Wiki}} \\
				\cmidrule(l){3-4} \cmidrule(l){5-6} 
				\textbf{Model}        & \textbf{Variants}                                     & top-1            & top-5             & top-1          & top-5 \\
				\midrule
				BraVL                 & w/o $\mathcal{L}_{intra}$, $\mathcal{L}_{inter}$      & 34.25$_{*}$             & 69.83$_{*}$              & 11.46$_{*}$           & 49.39$_{*}$               \\
				BraVL                 & w/o $\mathcal{L}_{intra}$                             & 35.03$_{*}$             & 70.43$_{*}$              & 12.30$_{*}$           & 51.44$_{*}$               \\
				BraVL                 & w/o $\mathcal{L}_{inter}$                             & 35.15$_{*}$             & 70.22$_{*}$              & 12.67$_{*}$           & 51.00$_{*}$               \\
				BraVL (ours)          & completed                                             & {\bftab 36.50}   & {\bftab 72.39}    & {\bftab 13.39} & {\bftab 53.54}              \\			
				\bottomrule
			\end{tabular}
			\vspace{-5pt}
		}
		\label{tbl:Ablation}
	\end{table}

	The choice of joint posterior approximation function directly influences the properties of our BraVL model.
	As shown in Table \ref{Table:posterior}, three choices can be followed for sampling the latent variable $\z$, i.e., the PoE \cite{MVAE_wu_2018}, MoE \cite{shi2019variational} and MoPoE \cite{sutter2021generalized}. 
	The ablation results are shown in Table \ref{tbl:posterior}, where we see that MoPoE results in significantly
	higher decoding accuracy than PoE and MoE. This also shows that MoPoE is better suited to the proposed multimodal learning method.
	\begin{table}[!htbp]     \centering
		\ra{0.9}
		\co{4.0pt}
		\vspace{-9pt}
		\caption{Ablation results of joint posterior approximation. * indicates that MoPoE is significantly better than the compared one (paired t-test, p$<$0.05).}
		\vspace{-6pt}
		{\small
			\begin{tabular}{@{}ll*{6}{l}@{}}
				\toprule
				& & \multicolumn{2}{c}{\makecell{DIR-Wiki}} & \multicolumn{2}{c}{\makecell{GOD-Wiki}} \\
				\cmidrule(l){3-4} \cmidrule(l){5-6}  
				\textbf{Model}        & \textbf{Posterior type}                   & top-1             & top-5                    & top-1                     & top-5 \\
				\midrule
				BraVL                 & PoE\cite{MVAE_wu_2018}                 & 31.42$_{*}$             & 66.87$_{*}$                    & 10.30$_{*}$                      & 39.56$_{*}$             \\
				BraVL                 & MoE\cite{shi2019variational}           & 35.71$_{*}$             & 72.33                    & 10.97$_{*}$                      & 48.17$_{*}$              \\
				BraVL (ours)          & MoPoE\cite{sutter2021generalized}      & {\bftab 36.50}    & {\bftab 72.39}           & {\bftab 13.39}           & {\bftab 53.54}             \\
				\bottomrule
			\end{tabular}
			\vspace{-5pt}
		}
		\label{tbl:posterior}
	\end{table}

	\subsubsection{Sensitivity analysis.}
\begin{figure*}[!htbp]
		\centering
		\includegraphics[scale=0.17]{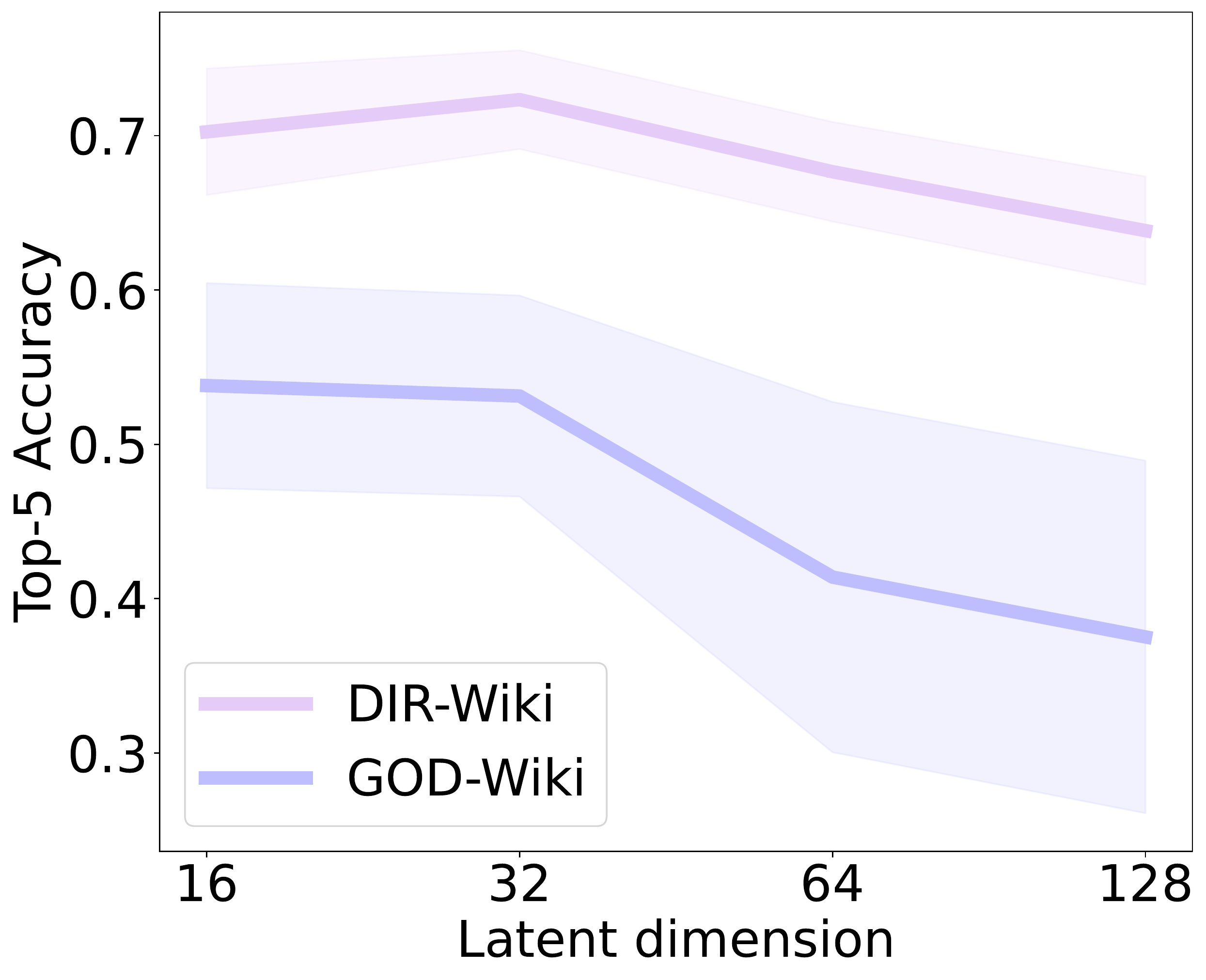}
		\hspace{2pt}
		\includegraphics[scale=0.17]{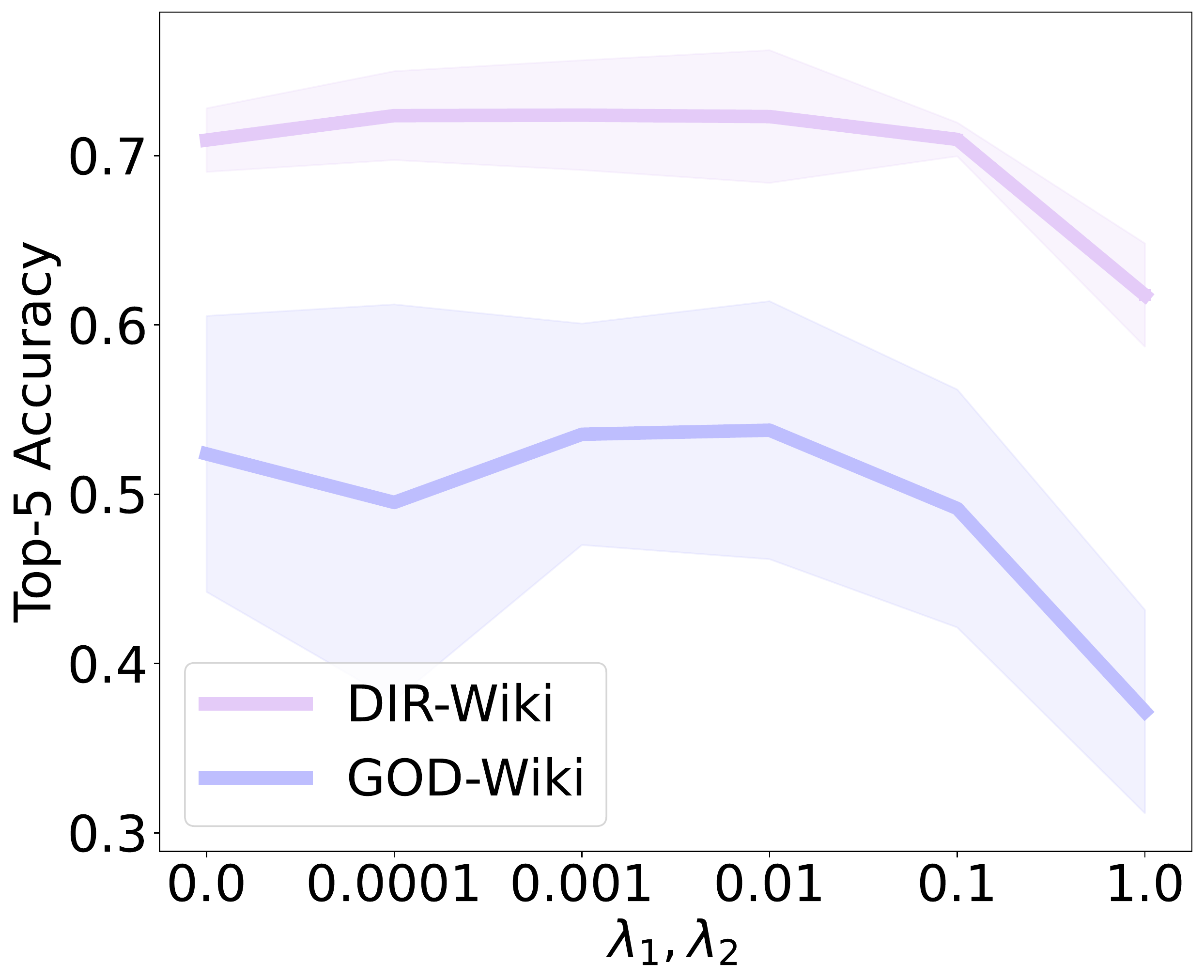}
		\hspace{2pt}
		\includegraphics[scale=0.17]{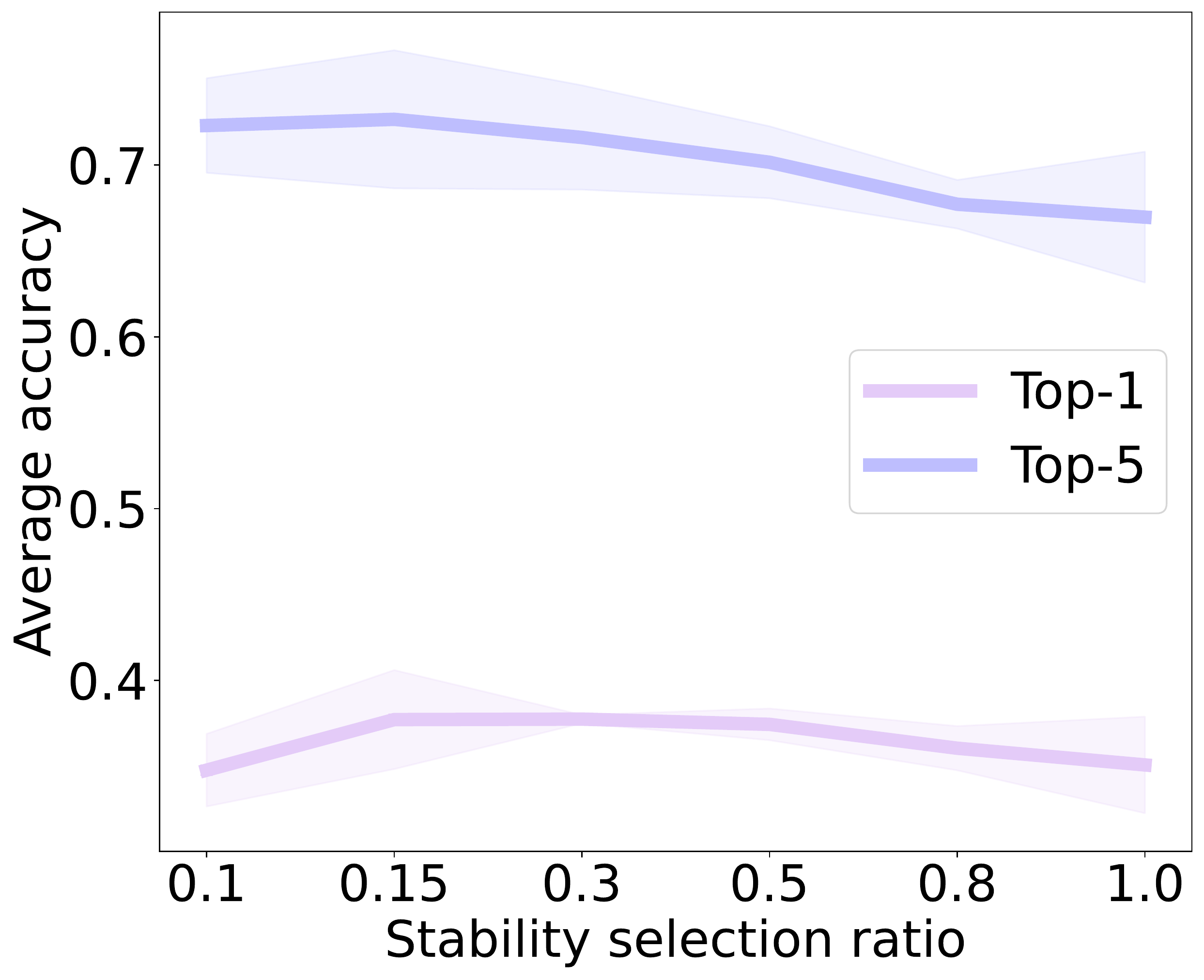}
		\hspace{2pt}
		\includegraphics[scale=0.17]{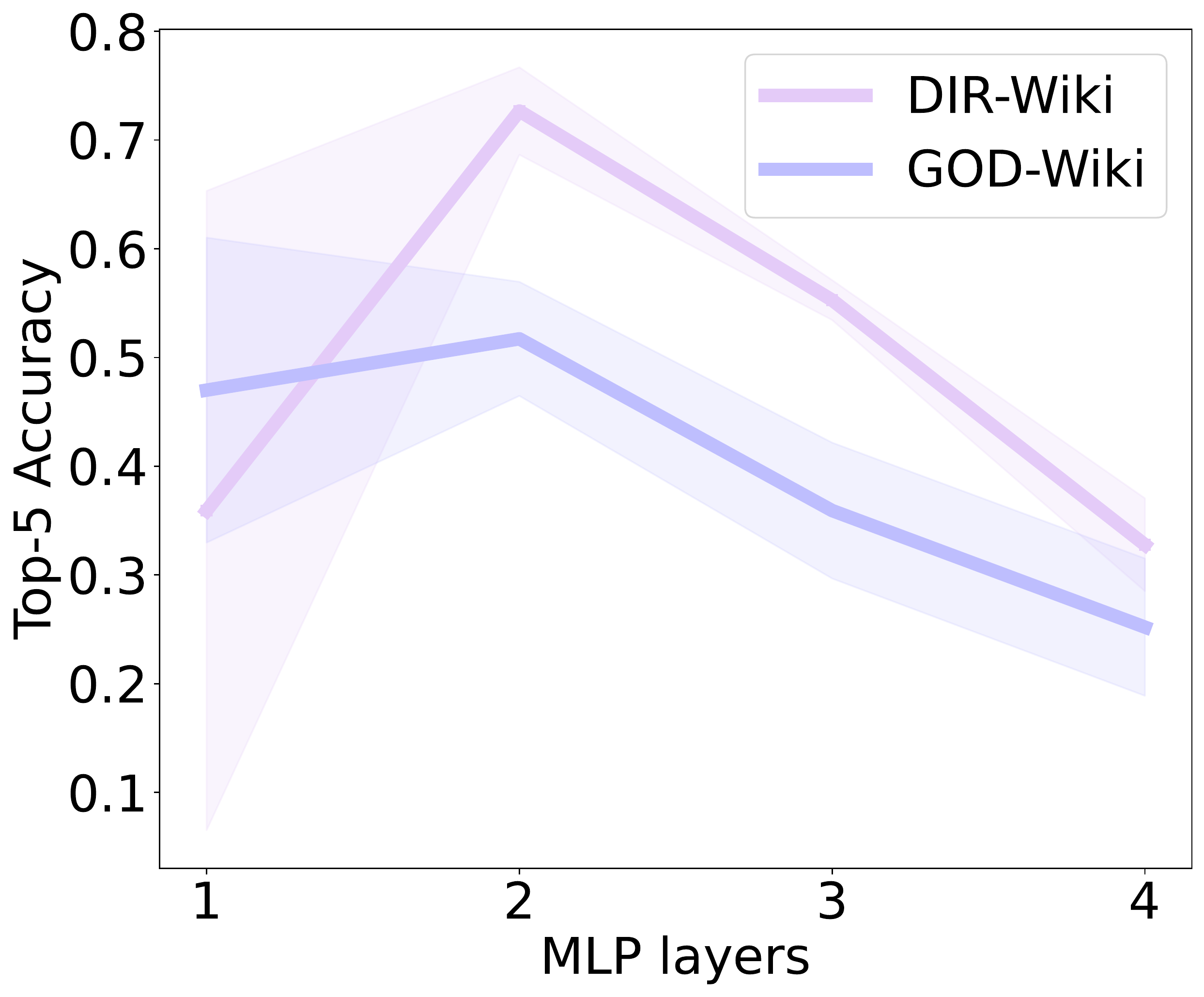}
		\vspace{-7pt}
		\caption{The impact of the latent dimension,  MI regularization coefficients $\lambda_1$ and $\lambda_2$, stability selection ratio, as well as the number of MLP layers.} 
		\vspace{-7pt}
		\label{fig:latent}
	\end{figure*}

	We also explored the robustness of our BraVL model to some hyper-parameters such as the latent space dimension, the MI regularization coefficients $\lambda_1$ and $\lambda_2$, the stability selection ratio, and the number of MLP layers in all encoders and decoders. Higher latent dimensions allow more degrees of freedom but require more data, while compact features capture the essential discriminative information. Without loss of generality, Fig. \ref{fig:latent} reports the averaged decoding accuracy across all subjects for different latent dimensions (16, 32, 64 and 128), $\lambda_1$ and $\lambda_2$, stability selection ratios, and MLP layers. We observe that the accuracy initially increases (or at least remains stable) with increasing latent dimension until it is equal to 32, after which it declines upon further increase of the latent dimension. Therefore, we use 32 dimensional latent features throughout the paper.  On the other hand, we observe that large values of $\lambda_1$ and $\lambda_2$ dramatically degrade the model performance, while values that are too small will not play the regularization role. We empirically find that it is safe to choose $\lambda_1$ and $\lambda_2$ from 0.001 to 0.01 in the experiments. In this range, mutual information regularization always brings some performance improvement to the model. In addition, we observe that the model achieves peak performance when the stability selection ratio is 0.15 on the DIR-Wiki dataset. Finally, we find that BraVL achieves peak performance when using 2-layer MLPs, and more layers degrade the performance on both datasets.  We speculate that the reason is that the brain, visual, and textual features are already high-level representations, and hence deeper nonlinear transformations are prone to over-fitting.

	\subsubsection{Analyzing the impact of extra data.}
\begin{figure}[t!]
		\centering 
		\includegraphics[scale=0.44]{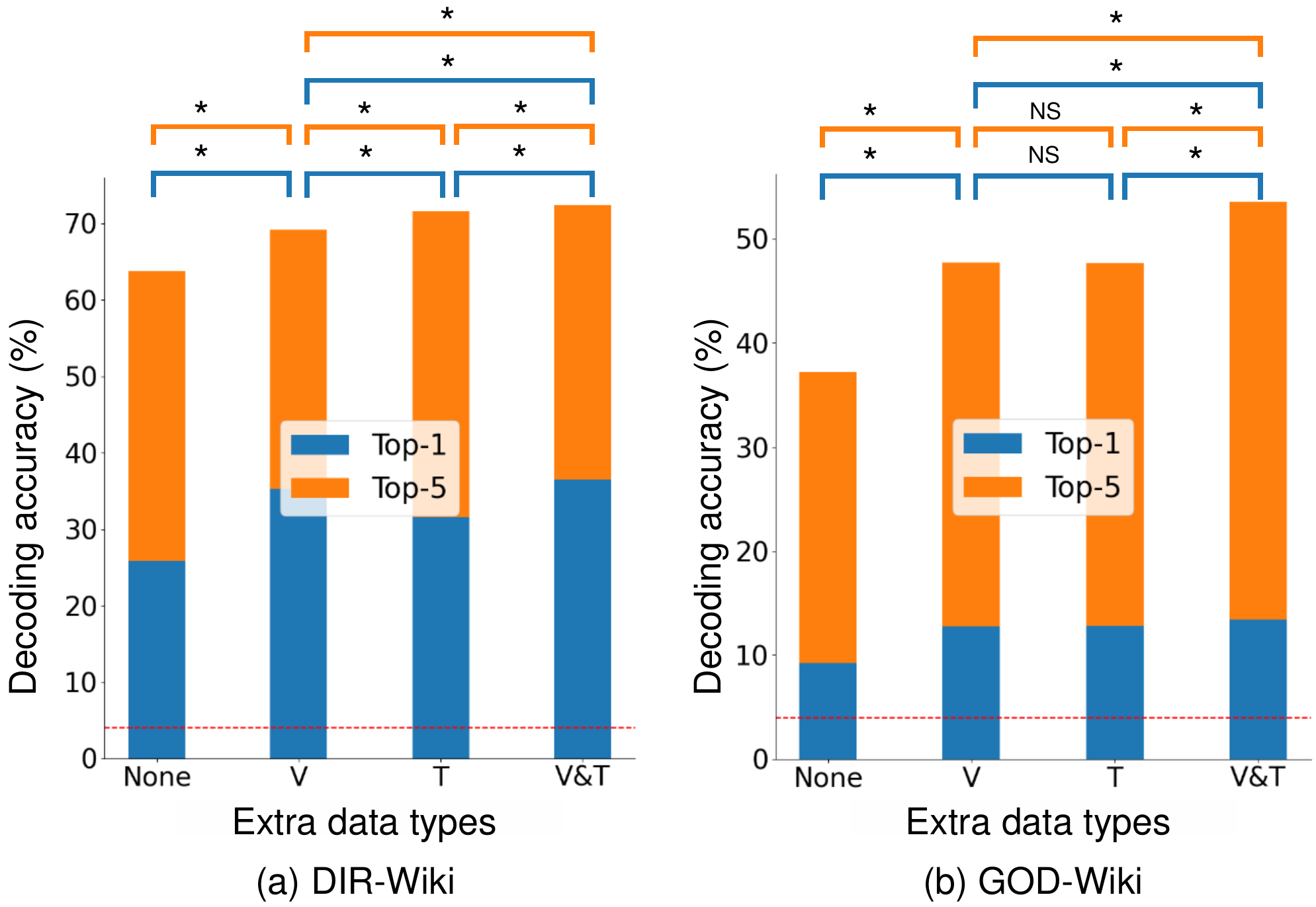}
		\vspace{-20pt}
		\caption{Decoding accuracy with different types of extra data. * indicates that the comparison is significant (paired t-test, p$<$0.05), and NS denotes that the comparison is not significant.  The red dashed lines represent the chance level.
		}
		\vspace{-6pt}
		\label{fig:extra_data_types}
	\end{figure}

\begin{figure}[t!]
		\centering 
		\includegraphics[scale=0.44]{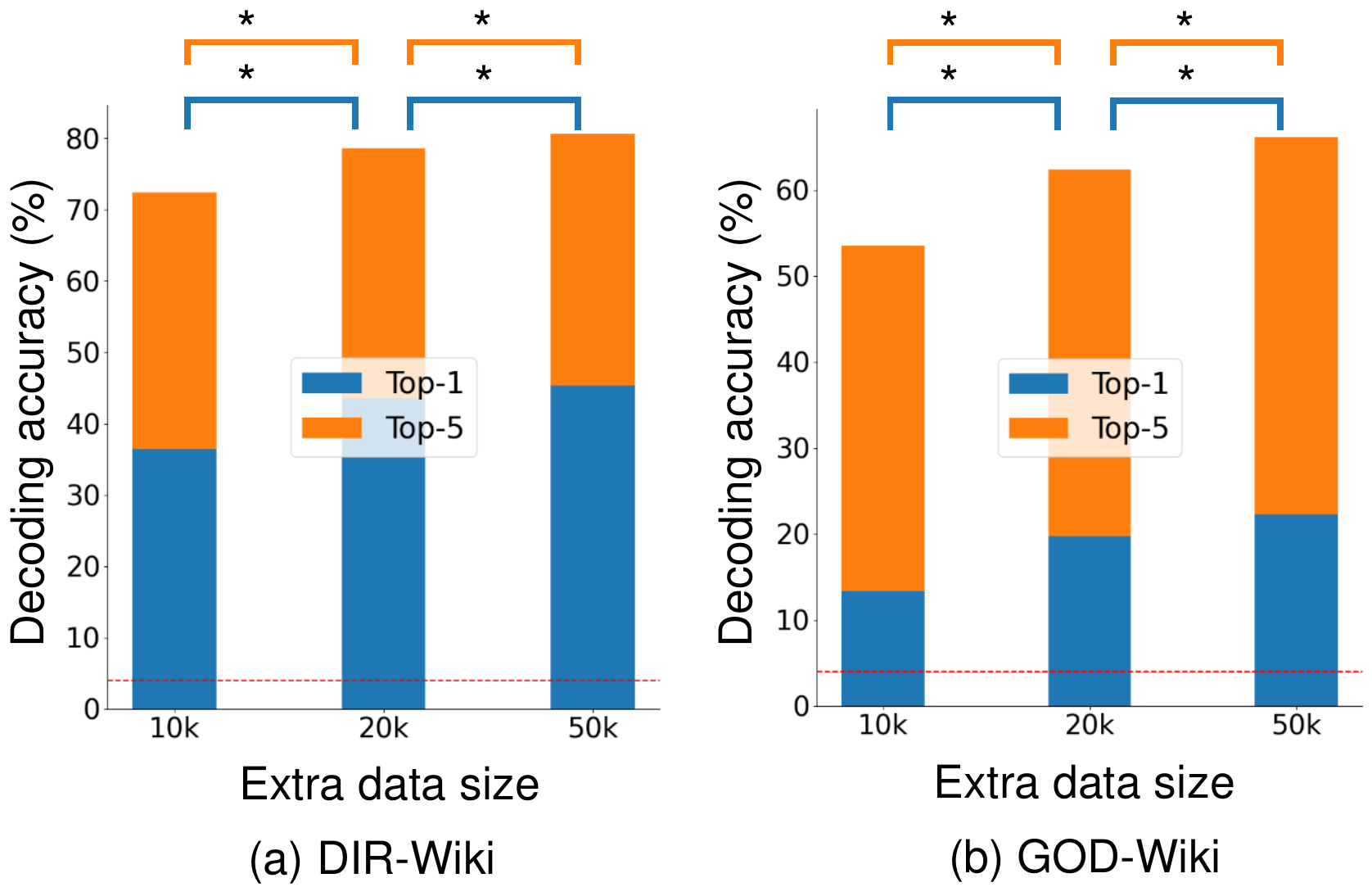}
		\vspace{-10pt}
		\caption{Decoding accuracy  with different numbers of extra V\&T data. * indicates that the comparison is significant (paired t-test, p$<$0.05). The red dashed lines represent the chance level.
		}
		\vspace{-13pt}
		\label{fig:extra_data_size}
	\end{figure}

	One advantage of our multimodal learning framework is that it can easily incorporate a large number of extra bimodal data (image-text pairs) or unimodal data (image/text) to facilitate joint representation learning. In Fig. \ref{fig:extra_data_types}, we analyzed the benefits of learning from different types of extra data for neural decoding.  We observe that using either unimodal extra data or bimodal extra data pairs can remarkably improve decoding accuracy. This suggests that the visual and textual features each provide reliable, complementary information to represent the semantic meaning of brain activity. Moreover, in Fig. \ref{fig:extra_data_size}, we also analyzed the effect of extra data size (10k, 20k and 50k, bimodal data) on neural decoding performance. The results show that the larger the extra dataset is, the better the decoding performance.

	\subsubsection{Cross-modality generation for brain activity.}
\begin{figure*}[!htbp]
		\centering
		\includegraphics[scale=0.82]{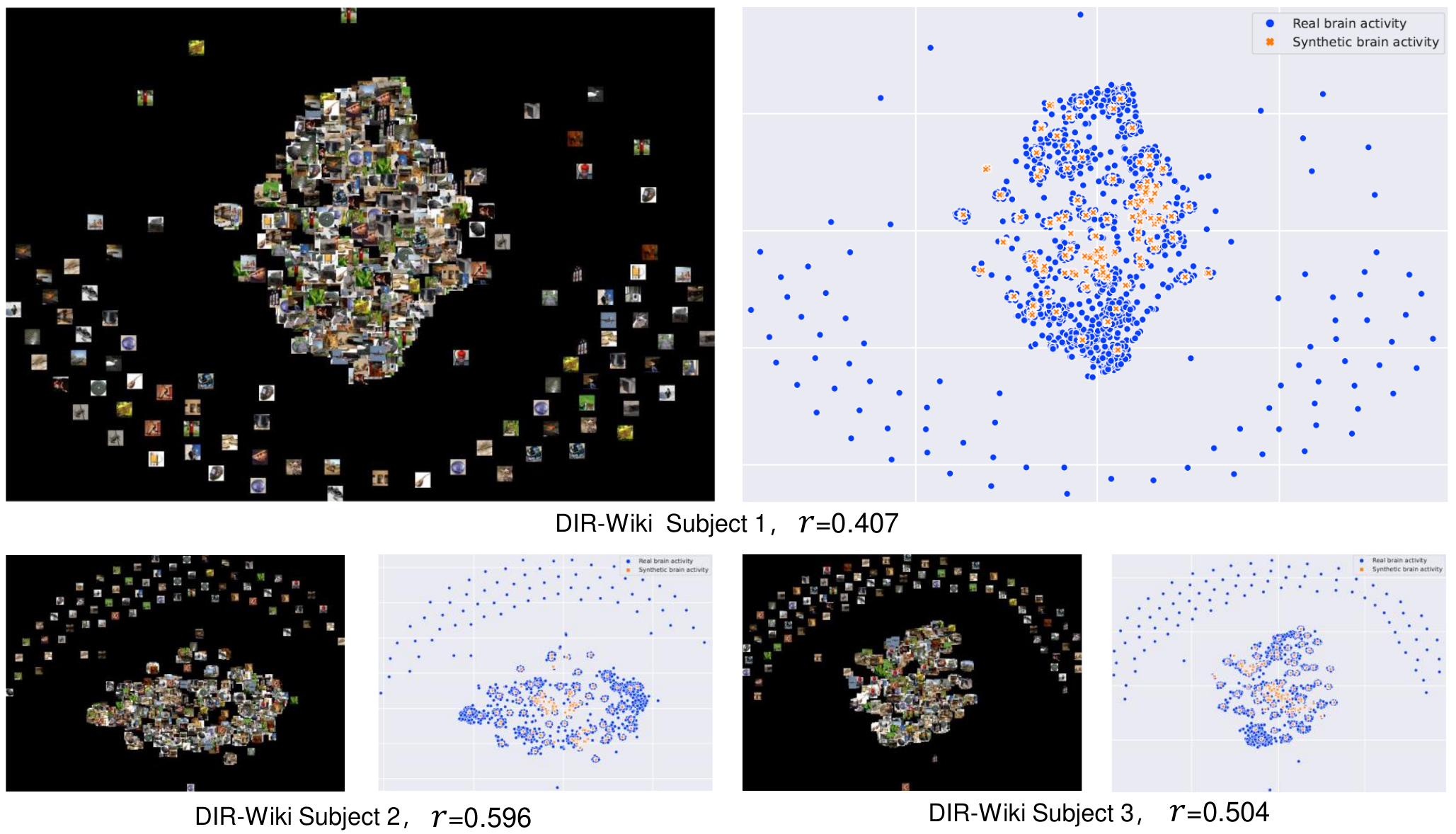}
		\includegraphics[scale=0.82]{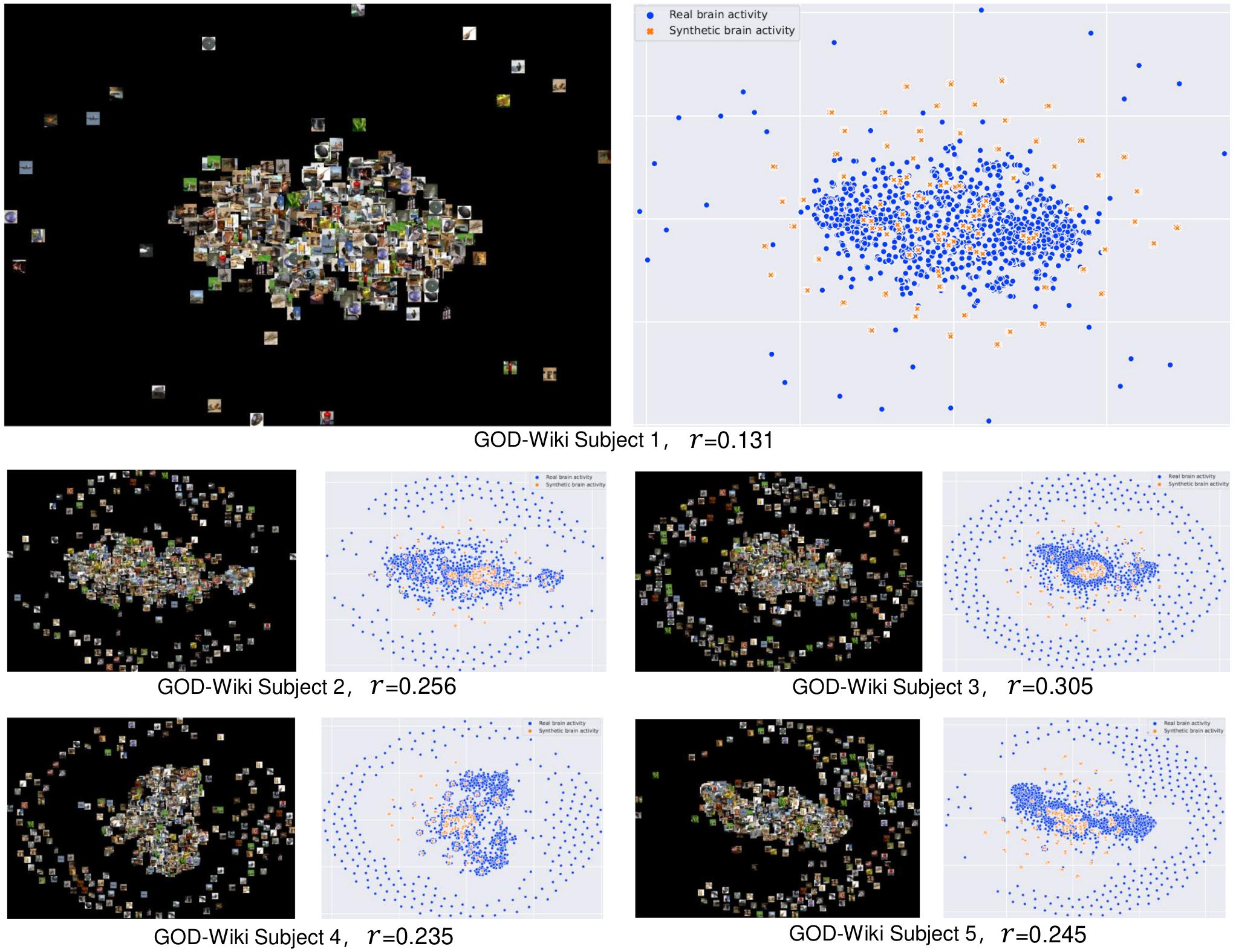}
		\vspace{-6pt}
		\caption{t-SNE visualization for the real (blue dot) and synthetic (orange cross) brain activity of novel classes (produced by the BraVL method, with V\&T features). To clearly understand which classes have better synthetic quality, we also put the corresponding images at the t-SNE space. Each point represents a testing sample (blue for real and orange for synthetic). The vertical axis and horizontal axis of each subfigure represent the two-dimensional spatial coordinates of the brain feature embedding fitted by t-SNE. The Pearson correlation coefficient (i.e., $r$) between the real and synthetic brain features is displayed below each subfigure.}
		\label{fig:synthetic}
	\end{figure*}

	Our BraVL model allows us to synthesize missing modalities in a cross-modality generation manner. Here, we treat the brain activity of novel classes as a missing modality and infer them from the visual and textual features of these novel classes.
	To examine the differences between synthetic and real brain activities of novel classes, we implement two kinds of comparisons: 1) visualization of data distribution (qualitatively) and 2) Pearson correlation coefficients (quantitatively).
	Due to the high dimensionality of brain activities, it is not convenient to visualize their distributions directly. Here, we use the t-SNE method \cite{van2008visualizing} to visualize them onto a two-dimensional plane, as shown in Fig. \ref{fig:synthetic}.
	To facilitate checking the visual category of each point in Fig. \ref{fig:synthetic}, we also put the corresponding images at the position. For each dimension of an individual subject's brain features, we measure the Pearson correlation coefficient between the sequences of real and synthetic brain responses to visual stimuli of novel classes, and display the average results below each subfigure.
	From Fig. \ref{fig:synthetic}, we observe that the synthetic brain activity and real brain activity have similar spatial data distributions, except for a few classes (due to a certain amount of noise in brain activity).   From the perspective of data distribution, these results can explain why our multimodal learning method can decode the neural representations of most novel classes. Moreover, from Fig. \ref{fig:synthetic}, we can also intuitively inspect which class cannot be successfully decoded and what about the individual differences between subjects.

	\subsubsection{Performance of different brain areas.} 
\begin{figure}[!htbp]
		\vskip -0.05in
		\centering\includegraphics[scale=0.17]{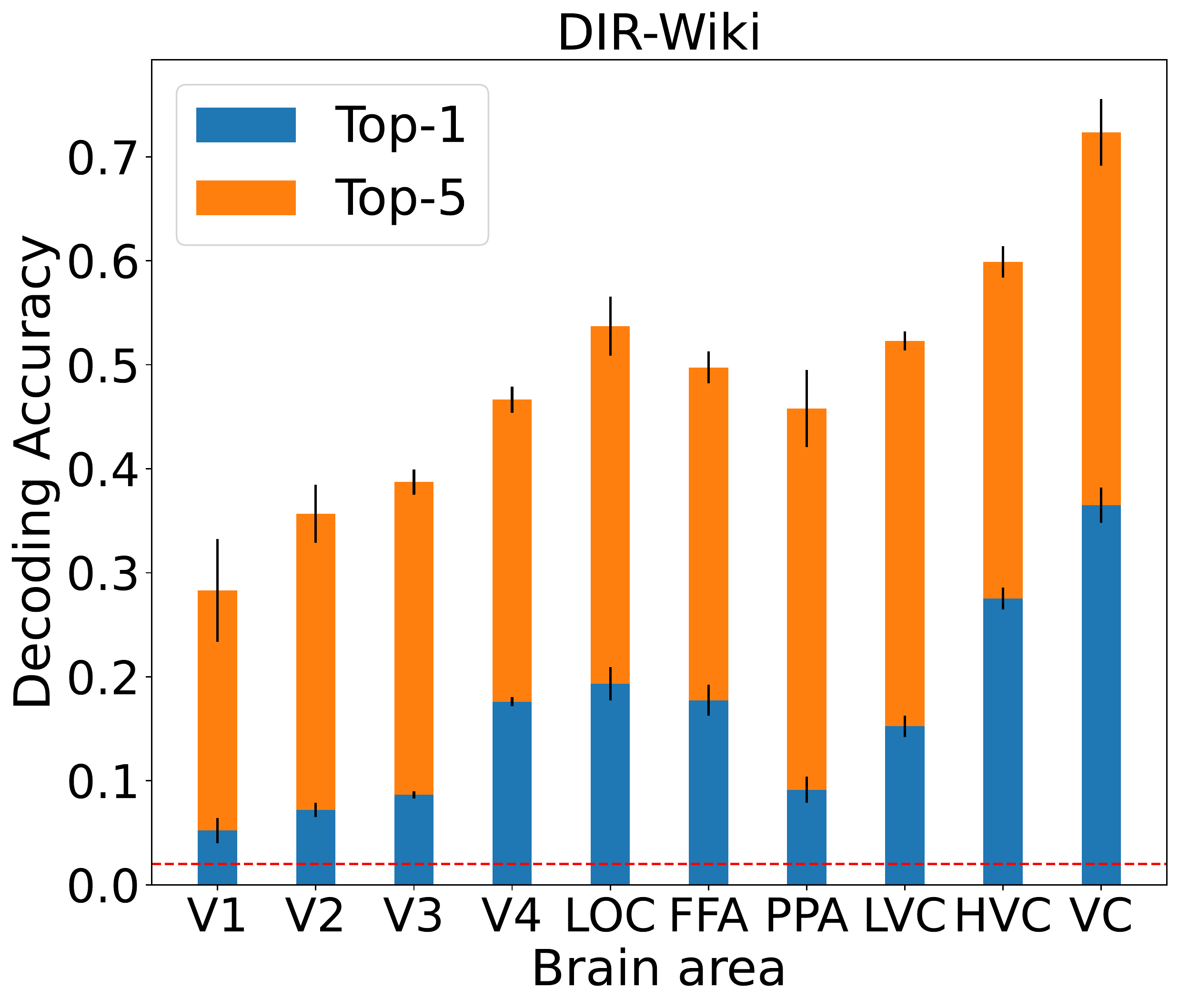}
		\centering\includegraphics[scale=0.17]{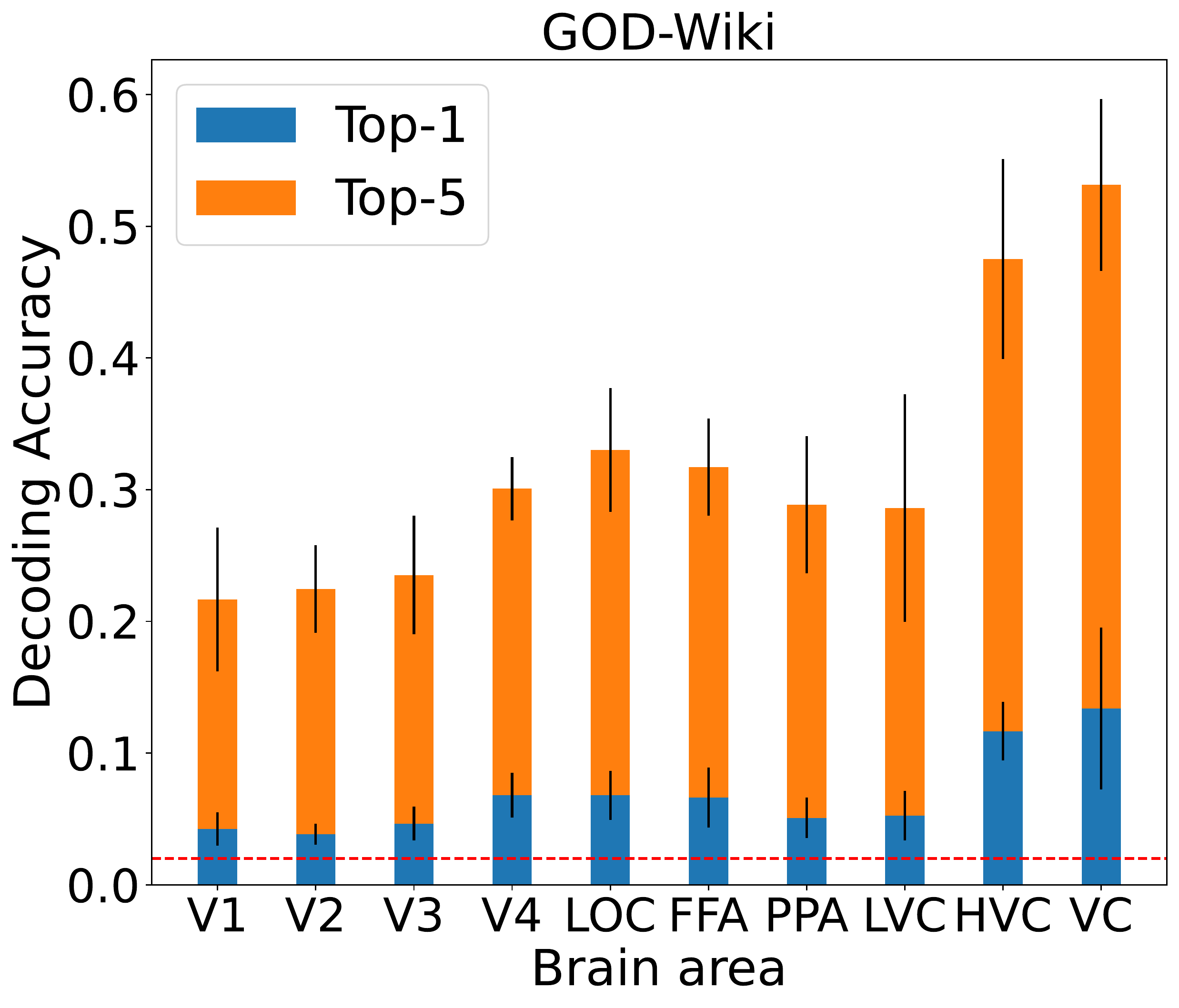}
		\vspace{-9pt}
		\caption{Decoding accuracy using different brain areas. The red dashed lines represent the chance level of Top-1 accuracy (0.02).}
		\vspace{-5pt}
		\label{fig:ROI}
	\end{figure}

	To validate the underlying brain areas that have a significant contribution to the decoding task, we separately run the experiments using fMRI voxels selected from different visual areas: V1, V2, V3, V4, LOC, FFA, PPA, etc.
	Voxels from regions V1-V3 are combined to form the LVC and voxels from LOC, FFA and PPA form the HVC. The whole visual cortex is denoted by `VC'. 
	Fig. \ref{fig:ROI} shows the decoding results produced by the proposed BraVL method using different brain areas.  We find that the decoding accuracy of HVC is better than that of LVC, and using the complete information of the entire VC always yielded the best results. 
	
\begin{figure}[!htbp]
		\centering 
		\includegraphics[scale=0.63]{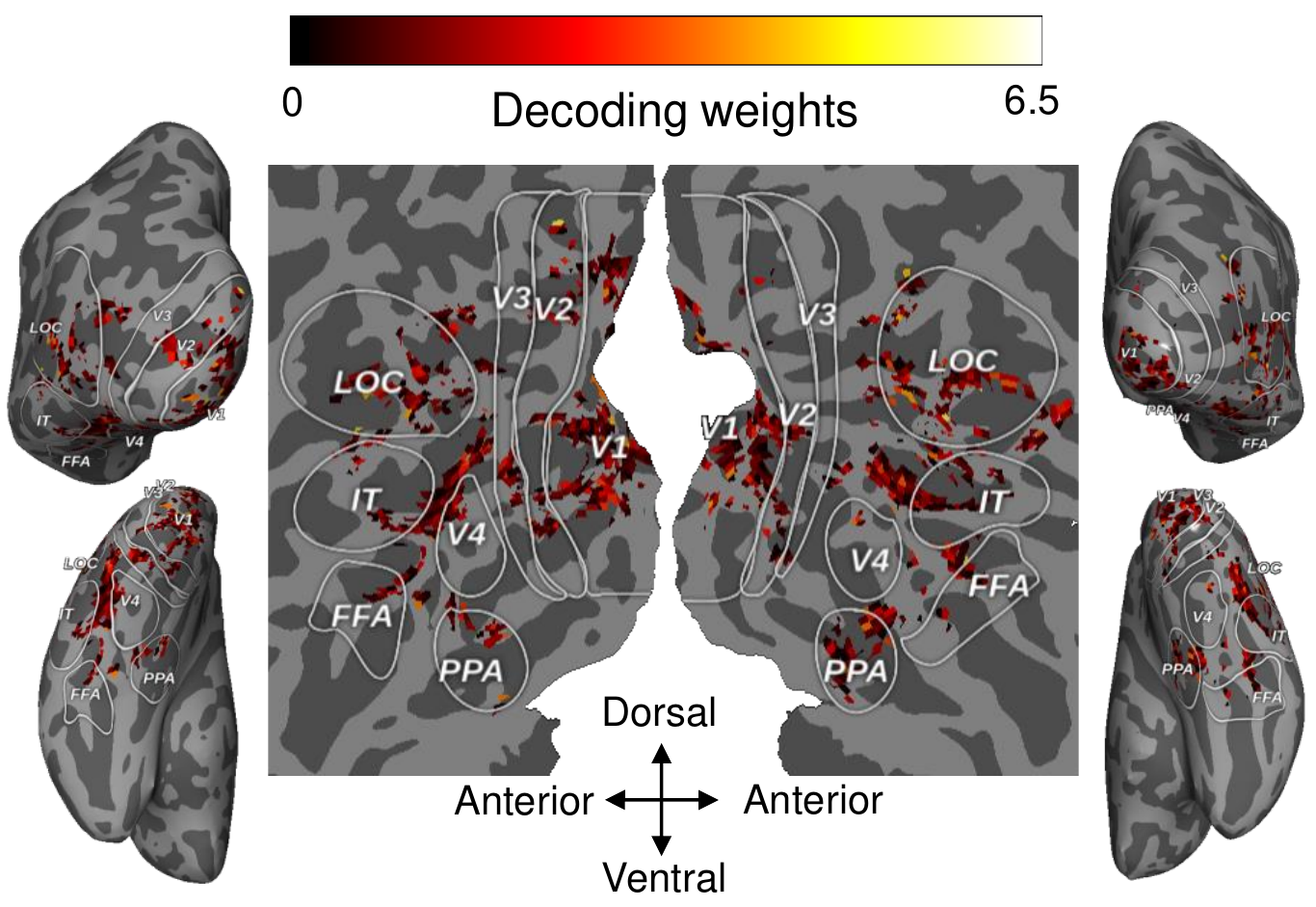}
		\vspace{-3pt}
		\caption{Visualization of the voxel-wise decoding weights onto the visual cortex (produced by the BraVL method, Subject 3, DIR-Wiki).}
		\vspace{-5pt}
		\label{fig:voxel_wise_importance}
	\end{figure}

	To visualize the contribution of each voxel to the decoding model, we first calculated the voxel-wise contribution weights $W$ using Eq.\ref{voxel_weights}
	\begin{equation}\label{voxel_weights}
		W =\bigg|\sum_i W_{PCA}^{l \times k} \times W_{fc}^{i}\bigg|,
	\end{equation}
	where $W_{PCA}$ represents the PCA projection matrix learned from the fMRI responses on the training set, $l$ denotes the number of voxels we used, $k$ denotes the dimension of the principal components, and $W_{fc}^{i}$ represents the fully connected weights for the $i$-th dimension of the first hidden layer of the brain Encoder. Subsequently, we projected the voxel-wise contribution weights onto the visual cortex, as illustrated in Fig. \ref{fig:voxel_wise_importance}. The weight assigned to each voxel was indicative of its contribution to brain decoding, with higher weights denoting greater importance relative to lower weights. The resulting analysis revealed a widespread distribution of importance weights across the visual cortex, suggesting that the visual cortex was extensively involved in the brain decoding process.

	\subsubsection{Evaluation on the ThingsEEG-Text dataset.}
	To evaluate the generalization ability of the proposed method on brain signals that are more convenient to collect in application, we also performed novel category neural decoding on the ThingsEEG-Text dataset. 
	All encoders and decoders on this dataset are MLPs with two hidden layers, each of which has 256 units. The BraVL model is trained for 500 epochs with a batch size of 1024, and the SVM classifier parameter '$gamma$'=1e-3. Other settings were identical to those on the fMRI datasets. The intra-subject and inter-subject decoding results are shown in Table \ref{tbl:comparision_eeg}. First, we again see that using the combination of visual and textual features (i.e., V\&T) performs significantly better than those using either of them alone (p$<$0.05), which is consistent with the results on the fMRI-based datasets (in Table \ref{tbl:comparision}). Second, BraVL trained with V\&T features achieves an average top-5 accuracy of 40.28\% and clearly outperforms the chance level (10\%). In particular, when the number of novel candidate classes is extended to 200, the average top-5 accuracy can still reach 17.45\%, which is much higher than the chance level of 2.5\%. Third, inter-subject neural decoding yields reasonable accuracies moderately lower than those obtained by intra-subject predictions but still much higher than the chance level. Together, these results provide evidence that the proposed model generalizes well on the EEG dataset, although the accuracy is lower than that on the fMRI datasets. 
	
	\begin{table*}[!htbp]     \centering
		\ra{0.9}
		\co{2.5pt}
		\caption{Intra-subject and inter-subject decoding accuracy (\%) of the proposed BraVL method trained on the ThingsEEG-Wiki dataset with the visual (V), textual (T) and combined (V\&T) features, respectively. $50/200$-way decoding means predicting the correct class out of $50/200$ novel classes. V$_{ft}$ means the visual feature extractor (i.e., CORnet-S) was fine-tuned on the training images of ThingsEEG-Text with 1654 classes.}
		\vspace{-7pt}
		\resizebox{18.3cm}{!}{
			\begin{tabular}{@{}lc*{24}{l}@{}}
				\toprule
				& & \multicolumn{24}{c}{\makecell{\textbf{Intra-subject: train and test the decoding models with data from the same subject}}}  \\
				\cmidrule(l){3-24} 
				& & \multicolumn{2}{l}{\makecell{Subject 1}} & \multicolumn{2}{l}{\makecell{Subject 2}} & \multicolumn{2}{l}{\makecell{Subject 3}} & \multicolumn{2}{l}{\makecell{Subject 4}} & \multicolumn{2}{l}{\makecell{Subject 5}}  & \multicolumn{2}{l}{\makecell{Subject 6}} & \multicolumn{2}{l}{\makecell{Subject 7}} & \multicolumn{2}{l}{\makecell{Subject 8}} & \multicolumn{2}{l}{\makecell{Subject 9}} & \multicolumn{2}{l}{\makecell{Subject 10}} & \multicolumn{2}{l}{\makecell{\textbf{Average}}} \\
				\cmidrule(l){3-4} \cmidrule(l){5-6} \cmidrule(l){7-8}  \cmidrule(l){9-10}  \cmidrule(l){11-12}  \cmidrule(l){13-14} \cmidrule(l){15-16}   \cmidrule(l){17-18}  \cmidrule(l){19-20} \cmidrule(l){21-22}  \cmidrule(l){23-24}
				\textbf{Type}                            & \textbf{Modality}  & top-1          & top-5          & top-1          & top-5     & top-1 & top-5  & top-1 & top-5   & top-1 & top-5  & top-1 & top-5  & top-1 & top-5    & top-1& top-5 & top-1& top-5 & top-1 & top-5 & top-1 & top-5\\
				\midrule
				& V  & 12.6 & 37.57 & 11.07 & 35.55 & 12.93 & 39.4 & 10.95 & 33.58 & 10.27 & 31.6 & 13.65 & 39.0 & 14.03 & 40.15 & 17.65 & 45.32 & 9.22 & 30.83 & 14.97 & 41.95 & 12.74 & 37.49   \\
				& V$_{ft}$  & 10.27 & 31.8 & 9.57 & 31.7 & 9.0 & 30.7 & 7.3 & 26.02 & 8.58 & 28.0 & 10.65 & 34.98 & 9.18 & 31.0 & 15.02 & 40.27 & 7.58 & 27.43 & 11.6 & 35.68 & 9.87 & 31.76   \\
				50-way                        & T   & 2.95 & 15.6 & 3.7 & 15.57 & 2.8 & 13.55 & 5.15 & 19.88 & 2.75 & 14.7 & 4.55 & 18.1 & 2.9 & 16.4 & 4.3 & 17.1 & 3.57 & 15.0 & 3.82 & 17.97 & 3.65 & 16.39   \\
				& V\&T & {\bftab 14.8} & {\bftab 41.5} & {\bftab 12.88} & {\bftab 39.15} & {\bftab 15.0} & {\bftab 40.85} & {\bftab 12.35} & {\bftab 36.45} & {\bftab 10.45} & {\bftab 33.77} & {\bftab 15.1} & {\bftab 41.17} & {\bftab 15.12} & {\bftab 42.38} & {\bftab 20.32} & {\bftab 49.83} & {\bftab 10.55} & {\bftab 34.1} & {\bftab 16.75} & {\bftab 43.6} & {\bftab 14.33} & {\bftab 40.28}    \\
				& V$_{ft}$\&T & 10.72 & 36.35 & 10.2 & 31.9 & 12.43 & 35.8 & 9.25 & 29.12 & 8.38 & 29.83 & 12.35 & 38.52 & 11.5 & 35.77 & 16.68 & 45.35 & 7.32 & 27.93 & 13.83 & 38.67 & 11.26 & 34.93    \\
				\midrule
				& V   & 5.42 & 16.29 & 3.81 & 13.19 & 5.29 & 15.76 & 4.37 & 13.64 & 3.54 & 12.32 & 5.46 & 16.86 & 6.06 & 17.21 & 7.36 & 21.02 & 3.94 & 12.66 & 6.04 & 17.62 & 5.13 & 15.66   \\
				& V$_{ft}$   & 3.67 & 12.28 & 3.23 & 11.47 & 3.56 & 12.07 & 2.43 & 9.08 & 2.66 & 10.1 & 4.25 & 13.79 & 3.41 & 12.29 & 5.01 & 16.06 & 2.61 & 9.74 & 3.74 & 12.9 & 3.46 & 11.98  \\
				200-way                       & T   & 0.91 & 4.06 & 0.95 & 4.24 & 1.06 & 4.54 & 1.36 & 6.08 & 0.97 & 4.32 & 1.25 & 5.74 & 0.94 & 4.75 & 1.4 & 5.94 & 0.93 & 3.99 & 1.43 & 6.36 & 1.12 & 5.0   \\
				& V\&T & {\bftab 6.11} & {\bftab 17.89} & {\bftab 4.9} & {\bftab 14.87} & {\bftab 5.58} & {\bftab 17.38} & {\bftab 4.96} & {\bftab 15.11} & {\bftab 4.01} & {\bftab 13.39} & {\bftab 6.01} & {\bftab 18.18} & {\bftab 6.51} & {\bftab 20.35} & {\bftab 8.79} & {\bftab 23.68} & {\bftab 4.34} & {\bftab 13.98} & {\bftab 7.04} & {\bftab 19.71} & {\bftab 5.82} & {\bftab 17.45} \\
				& V$_{ft}$\&T  & 4.21 & 14.16 & 3.44 & 12.06 & 4.13 & 13.49 & 2.99 & 11.14 & 2.94 & 10.96 & 4.9 & 15.54 & 4.04 & 14.27 & 5.92 & 18.61 & 2.86 & 10.88 & 4.73 & 14.86 & 4.02 & 13.6   \\
				\toprule
				& & \multicolumn{24}{c}{\makecell{\textbf{Inter-subject: leaving one subject out for testing and train the decoding models on the averaged data of all other subjects}}}  \\
				\cmidrule(l){3-24} 
				& & \multicolumn{2}{l}{\makecell{Subject 1}} & \multicolumn{2}{l}{\makecell{Subject 2}} & \multicolumn{2}{l}{\makecell{Subject 3}} & \multicolumn{2}{l}{\makecell{Subject 4}} & \multicolumn{2}{l}{\makecell{Subject 5}}  & \multicolumn{2}{l}{\makecell{Subject 6}} & \multicolumn{2}{l}{\makecell{Subject 7}} & \multicolumn{2}{l}{\makecell{Subject 8}} & \multicolumn{2}{l}{\makecell{Subject 9}} & \multicolumn{2}{l}{\makecell{Subject 10}} & \multicolumn{2}{l}{\makecell{\textbf{Average}}} \\
				\cmidrule(l){3-4} \cmidrule(l){5-6} \cmidrule(l){7-8}  \cmidrule(l){9-10}  \cmidrule(l){11-12}  \cmidrule(l){13-14} \cmidrule(l){15-16}   \cmidrule(l){17-18}  \cmidrule(l){19-20} \cmidrule(l){21-22}  \cmidrule(l){23-24}
				\textbf{Type}                            & \textbf{Modality}  & top-1          & top-5          & top-1          & top-5     & top-1 & top-5  & top-1 & top-5   & top-1 & top-5  & top-1 & top-5  & top-1 & top-5    & top-1& top-5 & top-1& top-5 & top-1 & top-5 & top-1 & top-5\\
				\midrule
				& V  & 5.5 & 21.68 & 4.35 & 20.03 & {\bftab 4.15} & {\bftab 18.45} & {\bftab 5.92} & {\bftab 19.82} & 4.65 & 18.43 & 5.2 & 21.32 & 6.22 & 24.02 & 5.78 & {\bftab 24.15} & 3.35 & 16.78 & 5.03 & 21.38 & 5.02 & 20.6   \\
				50-way                        & T & 2.75 & 12.07 & 2.95 & 13.15 & 2.75 & 11.92 & 3.72 & 15.3 & 2.93 & 12.75 & 3.02 & 13.98 & 2.17 & 12.12 & 2.73 & 12.95 & 2.53 & 12.8 & 3.3 & 14.05 & 2.88 & 13.11    \\
				& V\&T  & {\bftab 6.38} & {\bftab 22.98} & {\bftab 4.98} & {\bftab 20.7} & 3.92 & 17.8 & 5.6 & 18.6 & {\bftab 4.67} & {\bftab 19.38} & {\bftab 5.65} & {\bftab 23.08} & {\bftab 6.25} & {\bftab 24.12} & {\bftab 6.02} & 23.9 & {\bftab 4.58} & {\bftab 18.7} & {\bftab 5.85} & {\bftab 22.8} & {\bftab 5.39} & {\bftab 21.2}  \\
				\midrule
				& V    & 2.02 & 7.77 & {\bftab 1.51} & 6.03 & {\bftab 1.52} & 5.7 & {\bftab 1.93} & {\bftab 6.92} & 1.37 & 5.63 & 1.75 & 7.20 &  2.13 & 7.88 & 2.11 & 7.51 & 1.44 & 5.95 & 2.13 & 8.29 & 1.79 & 6.88  \\
				200-way                       & T   & 0.8 & 3.7 & 0.74 & 3.43 & 0.7 & 3.48 & 1.03 & 4.23 & 0.76 & 3.59 & 0.88 & 3.82 & 0.81 & 3.71 & 0.81 & 3.82 & 0.73 & 3.43 & 1.15 & 4.54 & 0.84 & 3.78   \\
				& V\&T & {\bftab 2.3} & {\bftab 7.99} & 1.49 & {\bftab 6.32} & 1.39 & {\bftab 5.88} & 1.73 & 6.65 & {\bftab 1.54} & {\bftab 5.64} & {\bftab 1.76} & {\bftab 7.24} & {\bftab 2.14} & {\bftab 8.06} & {\bftab 2.19} & {\bftab 7.57} & {\bftab 1.55} & {\bftab 6.38} & {\bftab 2.3} & {\bftab 8.52} & {\bftab 1.84} & {\bftab 7.02} \\
				\bottomrule
			\end{tabular}
		}
		\label{tbl:comparision_eeg}
	\end{table*}
	
	\subsubsection{Fine-tuning the feature extractors.}
	Generally, using pre-trained feature extractors is the preferred choice for neural encoding and decoding \cite{horikawa2017generic,shen2019deep,schrimpf2021the}. Since the visual feature extractors we used are pre-trained on the ImageNet dataset and all images in GOD-Wiki and DIR-WiKi also came from ImageNet, further fine-tuning is not necessary for these two datasets. Here, we fine-tuned the visual feature extractor on the ThingsEEG-Text dataset (i.e., CORnet-S) since it is independent of the ImageNet dataset. We started by replacing the pre-trained CORnet-S's 1000-neurons output layer with a 1654-neurons layer, where each neuron corresponded to one of the 1654 semantic concepts. Then, we trained one such model using the 16,540 training images as inputs and the corresponding class label as outputs (Adam optimizer, leaning rate 1e-4, 30 epochs, batch size 256). Finally, we deployed the fine-tuned CORnet-S networks to extract the visual features of the training and test images and evaluated their prediction performance through the same decoding analyses described before. The results are shown in Table \ref{tbl:comparision_eeg}, from which we observe that fine-tuning the visual features harms the decoding performance. We speculate that this is caused by the small size of the images (1654 classes, 10 images per class) used for fine-tuning compared to the pre-training ones (1.28 million images on ImageNet). On the other hand, since the textual feature extractors have been pre-trained on very large-scale datasets (e.g., CLIP was trained with 400 million image-text pairs), we did not fine-tune them on our own datasets. 
	
	\section{Discussion}
	The central goal of neural decoding (also known as ``mind reading") is to determine how much mental information can be decoded from brain activity \cite{du2019brain}. 
	The potential applications of the proposed method are threefold. 1) As a neural semantic decoding tool, our method would play an important role in the development of new neuroprosthetic devices that operate by reading semantic information from the human brain. Although this kind of application is not mature yet, our method provides the technical foundation for it. 2) Through cross-modality generation for brain activity, our method can also be used as a neural encoding tool for investigating how visual and language features are expressed on the human cerebral cortex, revealing which brain regions have multimodal properties (i.e., sensitive to both visual and language features). 3) The neural decodability of the model representation can be considered as an indicator of the brain-like level of that model \cite{nonaka2021brain}. Therefore, our method can be used as a brain-like property assessment tool to test which model's (visual or linguistic) representations are closer to human brain activity, thereby inspiring researchers to design more brain-like computational models.
	
	Because it is very expensive to collect human brain activity for the various visual categories, we usually only have the brain activity of very limited visual categories. However, large amounts of image and text data without corresponding brain activity can also provide us with additional useful information. Our approach can make full use of all types of data (trimodal, bimodal and unimodal) to improve the generalization ability of neural decoding. The key to our approach is that we align the distributions learned from each modality to a shard latent space that contains the essential multimodal information associated with the novel classes. In the application, the input of our method is only brain signals, and no other data are needed, so it can be easily applied to most neural decoding scenarios.
	
	We have shown that the features we extracted from brain activity, visual images and text descriptions are valid for decoding neural representations.
	However, the extracted visual features may not accurately reflect all stages of human visual processing, and a set of better features will facilitate the tasks.
	Alternatively, a larger pretrained language model, such as GPT-3 \cite{brown2020language}, can be used to extract text features that are more capable of zero-shot generalization.
	Moreover, although Wikipedia articles contain rich visual information, this information can easily be obscured by a large number of non-visual sentences.
	Addressing this issue with visual sentence extraction \cite{kil2021revisiting} or collecting more accurate and enriched visual descriptions by ChatGPT\cite{wu2023visual} and GPT-4 \cite{openai2023gpt4} are interesting works.
	Furthermore, since voxels in close brain spatial locations are likely to have similar activation patterns, our voxel stability selection method may select a group of good quality but spatially redundant voxels. Finally, while this study considered a relatively large amount of trimodal matching data in comparison to related studies, a larger and more diverse dataset (including extra bimodal or unimodal data) may be beneficial. We leave these aspects for future work. 
	\section{Conclusion}
	In this paper, we studied the effectiveness of brain-visual-linguistic multimodal modeling for neural decoding. We proposed a novel multimodal learning model by designing the intra- and inter-modality mutual information regularization principles.  We constructed three trimodal matching datasets, and the experiments on them lead to interesting cognitive insights that visual perception is accompanied by language influences to represent semantic meanings of visual stimuli. We also found
	that the visual-linguistic representations obtained by using the mixture-of-product-of-experts and mutual information regularizers perform better than their counterparts. 
	Our results suggest that decoding novel visual categories from human brain activity is practically possible with good accuracy. 
	We believe that our research will be of great value to both artificial intelligence and cognitive neuroscience researchers.

	%
	
\begin{figure}[!htbp]
		\centering 
		\vspace{-10pt}
		\includegraphics[scale=0.5]{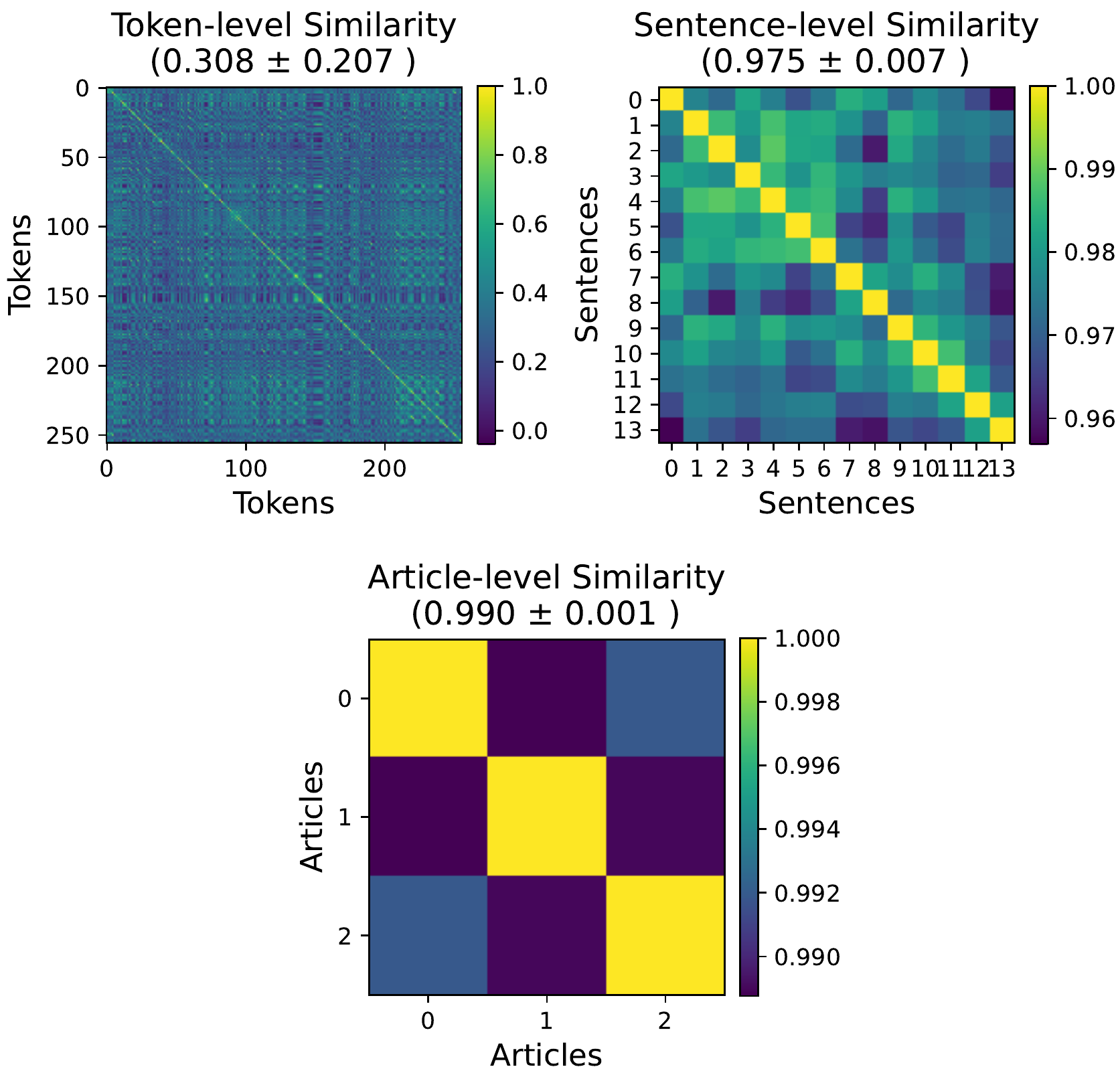}
		\vspace{-10pt}
		\caption{Cosine similarity matrices at different levels for Wiki. pages of ImageNet class n02817516 (`bearskin', `busby', `shako').}
		\vspace{-5pt}
		\label{fig:rsm}
	\end{figure}

	\appendix[Representational similarity measures]
	To show the degree of heterogeneity of text features before average pooling, for a typical example, we calculated the cosine similarity between each embedding pair (extracted by the ALBERT model~\cite{lan2019albert}) at the token-level, sentence-level and article-level, and the similarity matrices were shown in Fig. \ref{fig:rsm}. The mean$\pm$std was labeled in the title of each panel. Results for more examples can be found in Table \ref{Table:rsm}. 
	
	\begin{table}[!htbp]
		\co{3.0pt}
		\ra{1.0}
		\centering
		\caption{Cosine similarity measures (mean$\pm$std) for more examples.}
		\vspace{-6pt}
		\resizebox{8.2cm}{!}{
			\begin{tabular}{p{1.6cm}|p{1.7cm}|p{1.7cm}|p{1.7cm}|p{1.7cm}}
				\thickhline
				ImageNet class                             &Wiki. pages       &Token-level  &Sentence-level   & Article-level                \\ \hline
				`freight car' (n03393912)       &   `Goods wagon'                                      & \makecell*[l]{\hspace{-0.28em} \vspace{-1.5em} 0.289$\pm$0.197 }                     & \makecell*[l]{\hspace{-0.28em} \vspace{-1.5em} 0.966$\pm$0.018    }               & \makecell*[l]{\hspace{-0.28em} \vspace{-1.5em} 1.0$\pm$0.0   }          \\ \hline
				`marimba' (n03721384)       &  `Xylophone', `Marimba'                                      & \makecell*[l]{\hspace{-0.28em} \vspace{-1.5em} 0.297$\pm$0.199 }                     & \makecell*[l]{\hspace{-0.28em} \vspace{-1.5em} 0.960$\pm$0.022 }                  & \makecell*[l]{\hspace{-0.28em} \vspace{-1.5em} 0.997$\pm$0.001 }                    \\ \hline			  
				`goblet' (n03443371)                    & `Chalice', `Stemware', `Wine glass'     & \quad \quad 0.296$\pm$0.189                      & \quad \quad 0.969$\pm$0.018                   & \quad \quad 0.968$\pm$0.016  \\              
				\thickhline
			\end{tabular}
		}
		\vspace{-6pt}
		\label{Table:rsm}
	\end{table}
	
	
	%
	
	\ifCLASSOPTIONcaptionsoff
	\newpage
	\fi
	
	\bibliographystyle{IEEEtran}
	\bibliography{changdedu}
	
	
	\begin{IEEEbiography}[{\includegraphics[width=1in,height=1.25in,clip,keepaspectratio]{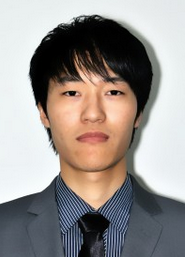}}]{Changde Du}
		received the Ph.D. degree from the Institute of Automation, Chinese Academy of Sciences (CASIA), in 2019.
		His current research interests include deep learning, computational neuroscience, brain-inspired intelligence, computer vision and brain-computer interfaces. He has published over 40 peer-reviewed research papers in prestigous conferences and journals. He won the following awards: National Scholarship for Doctoral Students (2018), President Prize of Chinese Academy of Sciences for Excellent Ph.D. Graduates (2019). His homepage: https://changdedu.github.io/.
	\end{IEEEbiography}

	\begin{IEEEbiography}[{\includegraphics[width=1in,height=1.25in,clip,keepaspectratio]{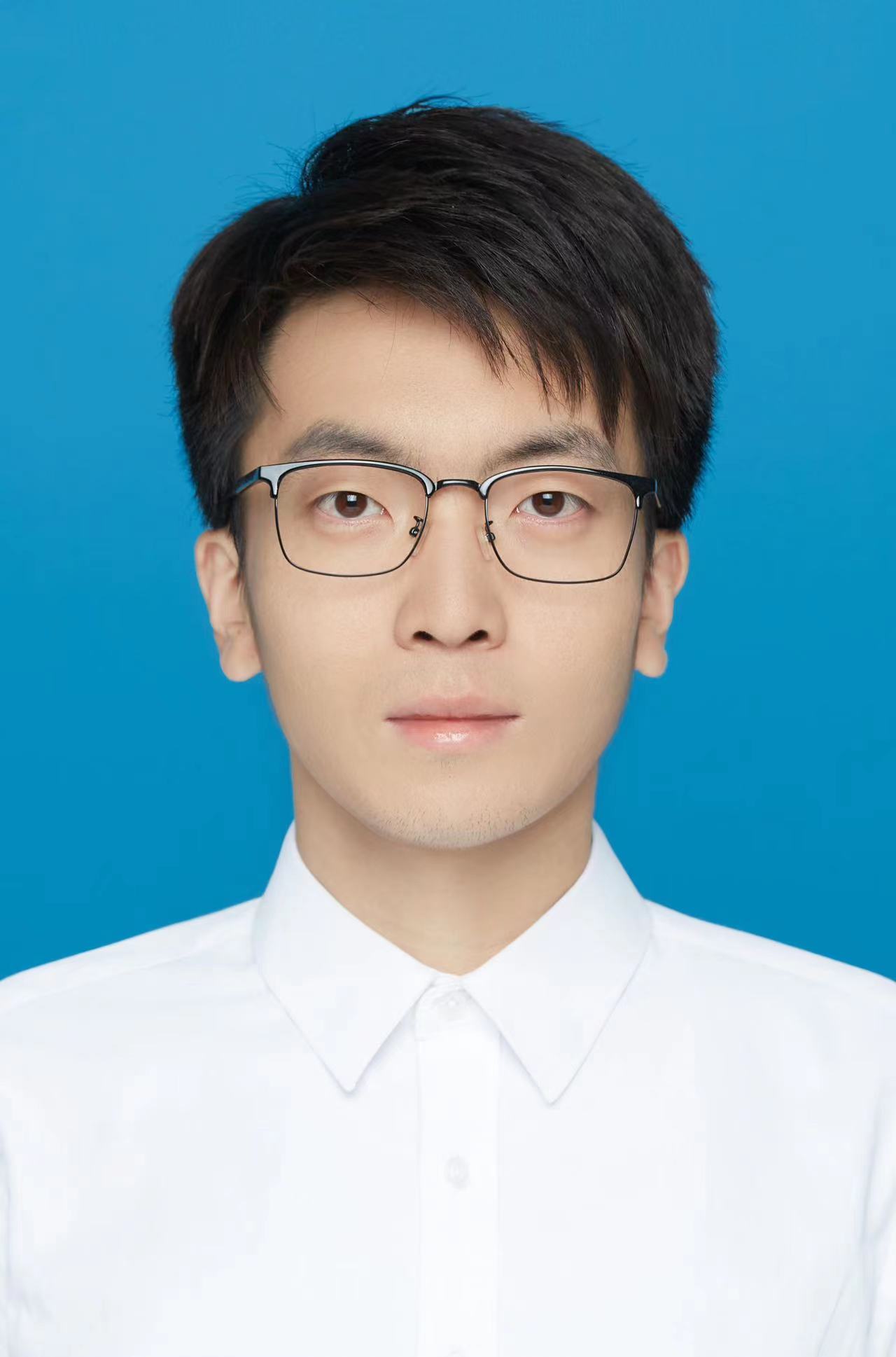}}]{Kaicheng Fu}
		received the B.S. degree in automation from Xi'an Jiaotong University, Xi'an, China, in 2019, and he is currently working toward a Ph.D. degree at University of Chinese Academy of Sciences, Beijing, China. His current research interests include emotion intelligence, multi-view learning and multi-label learning.
	\end{IEEEbiography}

	\begin{IEEEbiography}[{\includegraphics[width=1in,height=1.25in,clip,keepaspectratio]{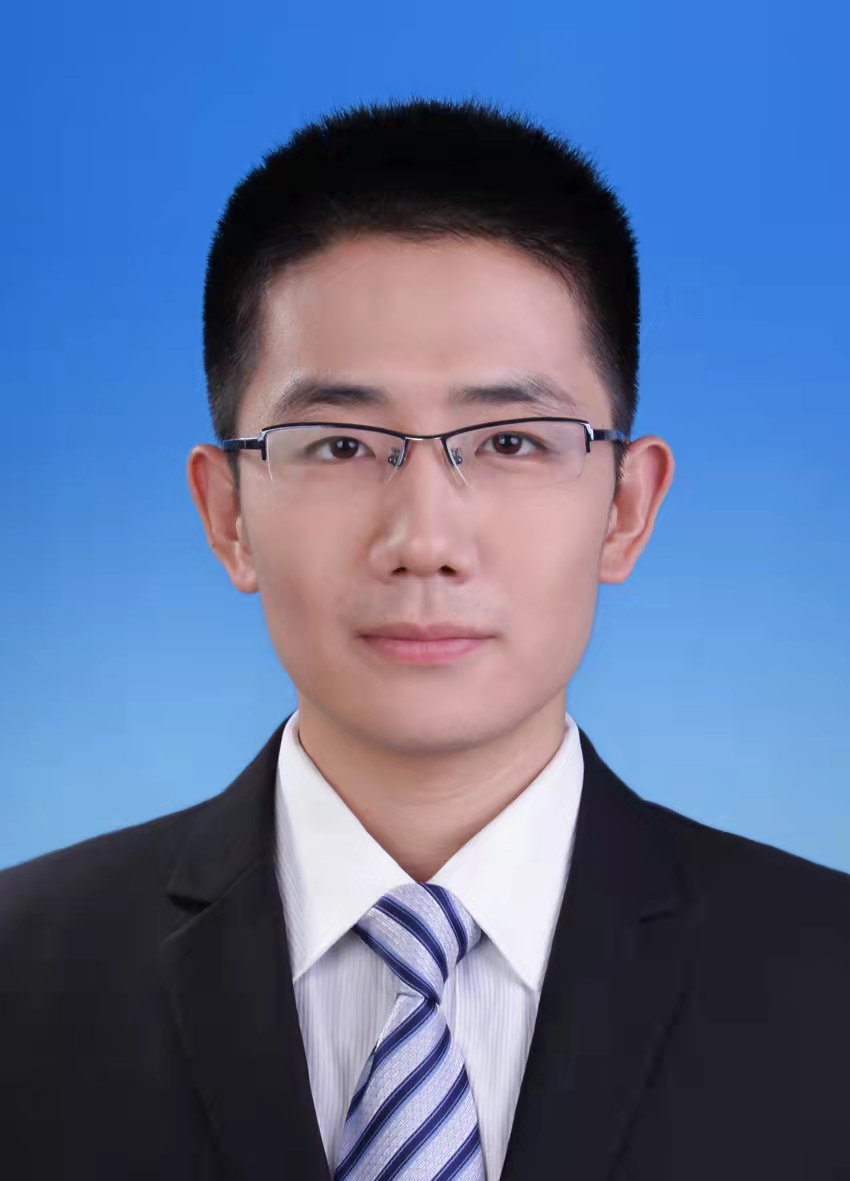}}]{Jinpeng Li}
		received the Ph.D. degree (2019) in Pattern Recognition and Intelligent System from National Laboratory of Pattern Recognition, Institute of Automation, Chinese Academy of Sciences. His research interests include machine learning, deep learning, transfer learning algorithms and their applications in brain-computer interfaces and medical image analysis.
	\end{IEEEbiography}

	\begin{IEEEbiography}[{\includegraphics[width=1in,height=1.25in,clip,keepaspectratio]{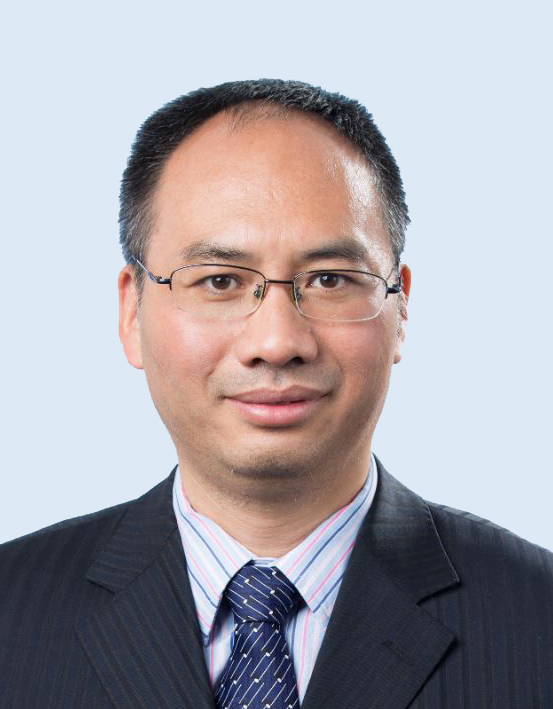}}]{Huiguang He}
		received the B.S. degree (1994) and the M.S. degree (1997) from Dalian Maritime University (DMU), Dalian, China. He got the Ph.D. degree (with honor) in pattern recognition and intelligent systems from Institute of Automation, Chinese Academy of Sciences (CASIA), Beijing, China. He is currently a full professor with CASIA. He was an Associate Lecturer in DMU from 1997 to 1999, and postdoctoral researcher in University of Rochester, USA from 2003 to 2004. He was a visiting professor in University of North Carolina at Chapel Hill from 2014 to 2015. He won the following awards: Excellent Ph.D. dissertation of CAS (2004), National Science \& Technology Award (2003, 2004), Beijing Science \& Technology Award (2002, 2003), K.C. Wong Education Prizes (2007, 2009),  “Jia-Xi Lu” Young Talent Prize (2009) and excellent member of Youth Innovation Promotion Association, CAS (2016). His research interests include pattern recognition, medical image processing, and brain computer interface (BCI). His research has been supported by several research grants from National Science Foundation of China, and he has published more than 200 peer-reviewed papers. He is a senior member of the IEEE.
	\end{IEEEbiography}
	
	
	

\end{document}